\documentclass{article}

\usepackage[preprint]{neurips_2024}

\usepackage{layouts}
\usepackage[utf8]{inputenc} % allow utf-8 input
\usepackage[T1]{fontenc}    % use 8-bit T1 fonts
\usepackage[hidelinks]{hyperref}       % hyperlinks
\usepackage{url}            % simple URL typesetting
\usepackage{booktabs}       % professional-quality tables
\usepackage{amsfonts}       % blackboard math symbols
\usepackage{nicefrac}       % compact symbols for 1/2, etc.
\usepackage{microtype}      % microtypography
\usepackage[table]{xcolor}         % colors
\usepackage{optidef}

\usepackage{booktabs}
\usepackage{mathtools}
\usepackage{amssymb}
\usepackage{multirow}
\usepackage{bm}
\usepackage{array}
\usepackage{multicol}
\usepackage{soul}
\usepackage{algorithm}
\usepackage{algorithmic}
\usepackage{xfrac}

\newcommand{\ttheta}{\boldsymbol{\theta}}

\title{OAC: Output-adaptive Calibration for Accurate Post-training Quantization}

\author{%
  Ali Edalati$^{1}$
  \quad
  Alireza Ghaffari$^{1,2}$
  \quad
  Mahsa Ghazvini Nejad$^{1}$
  \quad \\
  \textbf{Lu Hou}$^{1}$
  \quad
  \textbf{Boxing Chen}$^{1}$
  \quad
  \textbf{Masoud Asgharian}$^{2}$
  \quad
  \textbf{Vahid Partovi Nia}$^{1}$\thanks{Corresponding author: \texttt{vahid.partovinia@huawei.com}} \\
  $^{1}$Huawei Noah’s Ark Lab \\
  $^{2}$Department of Mathematics and Statistics, McGill University \\
}

\begin{document}

\maketitle

\begin{abstract}
Deployment of Large Language Models (LLMs) has major computational costs, due to their rapidly expanding size.  Compression of LLMs reduces the memory footprint, latency, and energy required for their inference. Post-training Quantization (PTQ) techniques have been developed to compress LLMs while avoiding expensive re-training. Most PTQ approaches formulate the quantization error based on a layer-wise Euclidean loss, ignoring the model output. Then, each layer is calibrated using its layer-wise Hessian to update the weights towards minimizing the quantization error. The Hessian is also used for detecting the most salient weights to quantization. Such PTQ approaches are prone to accuracy drop in low-precision quantization. We propose Output-adaptive Calibration (OAC) to incorporate the model output in the calibration process. We formulate the quantization error based on the distortion of the output cross-entropy loss. OAC approximates the output-adaptive Hessian for each layer under reasonable assumptions to reduce the computational complexity. The output-adaptive Hessians are used to update the weight matrices and detect the salient weights towards maintaining the model output. Our proposed method outperforms the state-of-the-art baselines such as SpQR and BiLLM, especially, at extreme low-precision (2-bit and binary) quantization.

\end{abstract}

\section{Introduction}
\begin{figure}[h!]
    \centering
    \includegraphics[width=\textwidth]{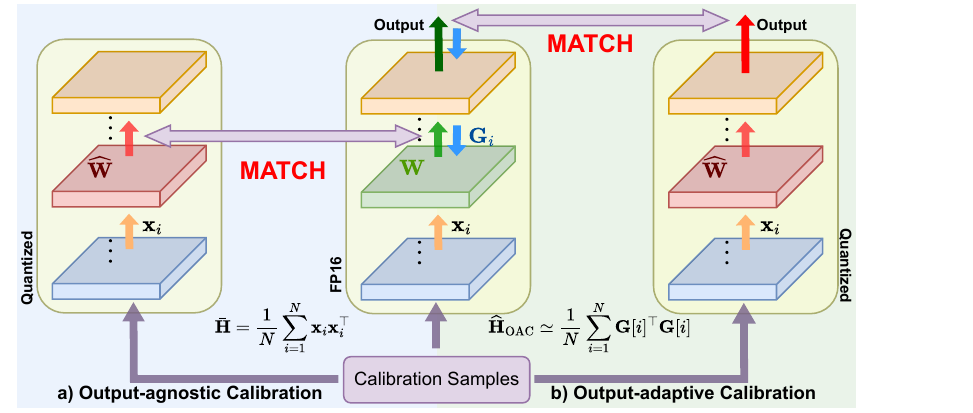}
    \caption{ a) Output-agnostic calibration minimizes the $\ell_2$ loss between the output of the quantized and original layers. b) Output-adaptive calibration matches the final output of the original and quantized models.}
    \label{fig:agnosticVSadaptive}
\end{figure}

The development of Large Language Models (LLMs) has sparked a revolution in natural language processing, leading to remarkable advancements in various tasks, including but not limited to reasoning, question answering, text generation, and few-shot learning \citep{unsupervised-learners,fewshot,opt,llama,llama2,gpt4}. LLMs comprise a huge number of parameters, exceeding tens and even hundreds of billions, which is considered one of the key factors to their success \citep{fewshot}. Their large size, however, is associated with major computational complexities and deployment barriers, especially, on resource-limited machines.

Post-training Quantization (PTQ) approaches have been introduced to overcome LLMs deployment challenges by reducing the precision of the model weights or activations while maintaining the performance without extra training \citep{smoothquant,zeroquant,optq,awq,squeezellm,dettmers2024spqr,billm}. PTQ techniques eliminate the necessity of Quantization-aware Training (QAT) methods \citep{qat1,llm-qat,shao2024omniquant,du2024bitdistiller} that are computationally demanding for LLMs. PTQ methods are one of the main techniques for efficient deployment of LLMs, enabling near loss-less compression up to 4 bits \citep{dettmers2024spqr}. However, extremely low-precision (2-bit or binary) PTQ is still an open challenge, which is the main focus of our proposed method.

To recover the performance of the quantized model, most novel PTQ methods \citep{optq,quip,dettmers2024spqr,billm} perform a layer-wise calibration to reduce the quantization error. These methods measure the quantization error by the $\ell_2$ loss between the output of the original and quantized layers over a small calibration set as shown in Figure \ref{fig:agnosticVSadaptive}:a). We refer to such approaches as \emph{output-agnostic calibration} because they only consider the output of the individual linear layers and ignore the model output. The output-agnostic methods compute the second derivative of the $\ell_2$ loss for the weights of each layer to formulate a layer-wise Hessian. The Hessian is used to measure the saliency of weights in order to detect the most salient ones as outliers \citep{dettmers2024spqr,squeezellm,billm}. Also, according to the Hessian, the weights are updated toward minimizing the $\ell_2$ error \citep{optq,quip,dettmers2024spqr,billm}. Given their output-agnostic nature, they are deemed inadequate to recover the performance of the original model at extreme low-precision quantization.

% they often fail to recover the performance of the original model at extreme low-precision quantization.  

We introduce \textbf{Output-adaptive Calibration (OAC)} to directly minimize the distortion of the model output after quantization, as shown in Figure \ref{fig:agnosticVSadaptive}:b). OAC reduces the quantization error of the model output and significantly outperforms the state-of-the-art PTQ methods at extreme low-precision quantization such as 2-bit and binary PTQ.

OAC follows the layer-by-layer recipe to quantize and calibrate the linear layers of LLMs. Our proposed method, however, employs the second derivative of the output cross-entropy loss to compute the \emph{output-adaptive Hessian} in contrast to the existing PTQ methods that use the Hessian of the layer-wise $\ell_2$ loss. We utilize the output-adaptive Hessian for updating the weights and measuring their saliency to accurately calibrate the model. The computation of the exact output-adaptive Hessian for large models is, however, computationally infeasible. To this end, we employ several techniques to reduce the computational complexity of approximating the output-adaptive Hessian.

Our proposed method, OAC, adopts the common assumptions in the PTQ literature \citep{obc,optq,dettmers2024spqr}, e.g. (i) the independence of linear layers, and (ii) the independence of the rows of the weight matrices. Assumption (i) is used for computation of the output-adaptive Hessian matrix of each layer, while assumption (ii) is used, in conjunction with the Fisher Information Identity, towards approximating the row-wise Hessians. Moreover, we formulate our quadratic optimization problem by aggregating the row-wise Hessian matrices to significantly reduce the memory footprint.

We apply our proposed method, OAC, for extreme low-precision PTQ of several LLMs and evaluate their performance on various tasks. OAC outperforms all of the state-of-the-art PTQ methods in 2-bit and binary PTQ of LLMs, by a significant margin. 

To summarize, our main contributions are:
\begin{itemize}
    \item We propose OAC, a novel output-adaptive calibration method for PTQ of LLMs that outperforms the state-of-the-art, especially in extreme low-precision quantization. OAC achieves superior results for binary and 2-bit quantization. To the best of our knowledge, this is the first output-adaptive calibration method for LLMs.
    
    \item We propose a quantization technique that minimizes the layer-wise quantization effect on the output cross-entropy loss. To achieve this, we develop an efficient Hessian approximation technique as the core component of our method. Integrating our proposed output-adaptive Hessian into other Hessian-based PTQ methods improves the accuracy.
    
    \item We provide theoretical insights on how to approximate the Hessian and provide a thorough experimental evaluation of various tasks to support our proposed method.
\end{itemize}

\section{Related Work}
There are two main categories of quantization techniques notably known as Quantization-aware Training (QAT) and Post-training Quantization (PTQ). QAT methods perform the quantization and training simultaneously \citep{llm-qat,Jacob_2018_CVPR,NIPS2017_1c303b0e,shao2024omniquant}. Considering the computational costs of re-training large models, QAT techniques are quite costly. The viable alternative is employing PTQ techniques, which utilize accurate solvers to minimize the quantization error without further training. Many PTQ methods leverage calibration that slightly modifies the model weights to reduce the quantization error on a small calibration set \citep{gholami2022survey}.

Among the existing  PTQ methods, AdaRound \citep{adaround}, OBQ \citep{obc}, AdaQuant \citep{adaquant}, and BRECQ \citep{brecq} are designed and applied to small computer vision models with around 100 million parameters. However, applying these methods to LLMs with billions of parameters is computationally intensive.

ZeroQuant \citep{zeroquant}, LLM.int8() \citep{llm-int8}, and SmoothQuant \citep{smoothquant} are among the first techniques that investigate PTQ of LLMs. ZeroQuant explores the quantization granularity in addition to layer-wise knowledge distillation. LLM.int8() separates the outlier activations using a threshold while quantizing the model to INT8 format. SmoothQuant \citep{smoothquant} reduces the activation quantization complexity by scaling the inputs of layers. Despite succeeding at 8-bit quantization of LLMs, ZeroQuant, SmoothQuant, and LLM.int8() perform poorly on extreme low-precision quantization such as 2-bit. 

Inspired by \citep{obs,obd}, OPTQ \citep{optq} performs a column-wise calibration procedure on the weights, which enables more accurate  3- and 4-bit PTQ of LLMs using a reasonable budget on time and computational resources. AWQ \citep{awq} detects the salient weights and scales them to reduce the PTQ error. QuIP \citep{quip} uses a generalized version of the OPTQ update formula in addition to incoherence pre-processing of the weights and Hessians to reduce the accuracy drop at sub-4-bit PTQ. QuIP\# \citep{quip2} enhances QuIP by randomized Hadamard transform \citep{hadamard}, vector quantization, and fine-tuning. SpQR \citep{dettmers2024spqr} improves the OPTQ calibration strategy with two major modifications to achieve near loss-less compression of LLMs up to 4-bits, (i) detecting and isolating the outliers by the Hessian (ii) performing a second round of quantization on the quantization parameters such as scales and zeros to reduce the average bit-width while keeping the outliers at FP32 format and using small group quantization.

SqueezeLLM \citep{squeezellm} investigates the non-uniform PTQ of LLMs using sensitivity-based K-means clustering without any calibration. To detect the salient weights, SqueezeLLM approximates the Hessian of each linear layer considering the output loss using the Fisher Information Identity. However, the non-uniform quantization is often associated with deployment challenges during the inference \citep{gholami2022survey}. OmniQuant \citep{shao2024omniquant} introduced an efficient QAT method that freezes the model weights while learning the quantization parameters to achieve the accuracy of QAT methods with a light training recipe. Orthogonal to our work, AQLM \citep{aqlm} uses Additive Quantization \citep{additive_quantization} as well as fine-tuning to overcome low-precision quantization. PB-LLM \citep{pbllm} investigated partial binarization of LLMs through a two-stage quantization recipe, (i) performing an OPTQ-based PTQ method on the non-salient binarized weights, (ii) performing QAT on the quantized model to further recover the accuracy while the salient weights are frozen. BiLLM \citep{billm} developed a more accurate PTQ method for the binarization of LLMs by identifying and structurally selecting salient weights. Furthermore, BiLLM minimizes the quantization error using a binary residual approximation strategy. BiLLM also employs splitting search to group and quantizes non-outlier weights based on their bell-shaped distribution.

Our proposed method has the following advantages compared to the state-of-the-art PTQ methods.
\begin{itemize}
    \item Our method computes the output-adaptive Hessian to calibrate the weights considering the model output cross-entropy loss, while OPTQ, QuIP, QuIP\#, SpQR, and BiLLM use the response-agnostic layer-wise Hessian.
    \item In contrast to SqueezeLLM and BRECQ, our method does not rely on the assumption that the output-adaptive Hessian is diagonal, and achieves a more accurate approximation. Moreover, BRECQ cannot be scaled to LLMs due to its computational complexity. Non-uniform quantization of SqueezeLLM is associated with deployment challenges and also fails to support extreme low-precision quantization.     
\end{itemize} 

\section{Output-agnostic Calibration Background}
\label{sec:agnostic}
In this section, we describe our notation as well as the response-agnostic calibration setting which is used by the status quo PTQ methods \citep{obc,optq,quip,dettmers2024spqr,billm}.

Consider an LLM, trained using the cross-entropy (CE) loss. Let $\ttheta \in \mathbb{R}^{D}$ be a $D$-dimensional vector of model weights comprising the weight matrices $\mathbf{W}^{(l)} \in \mathbb{R}^{d_{\mathrm{row}} \times d_{\mathrm{col}}}, l=1,\ldots, L$ i.e. $\ttheta=\mathrm{vec}[\mathbf W^{(l)}]_{l=1}^L$, where $L$ is the total number of layers. Let $\mathbf{y} \in \mathbb{R}^{d_\mathrm{vocab}} $ denote the model output given the input $\mathbf{x} \in \mathbb R^{d_\mathrm{col}}$. To simplify the notation, we drop $l$ from $\mathbf{W}$ when there is no risk of confusion. 
Also, $\mathbf{x}^{(l)} \in \mathbb{R}^{d_{\mathrm{col}}}$ denotes the input of the $l^{\mathrm{th}}$ layer. In what follows,  $\mathbf{A}_{:,q}$, $\mathbf{A}_{q,:}$, and $\mathbf{A}_{j,k}$ denote the $q^{\mathrm{th}}$ column, $q^{\mathrm{th}}$ row, and the $(j, k)^{\mathrm{th}}$ element of the matrix $\mathbf{A}$, respectively.  Most PTQ methods, follow a layer-by-layer recipe to quantize and calibrate the linear layers of LLMs.

Let $\mathbf{W}$ denote the weight matrix of the $l^{\mathrm{th}}$ linear layer. The quantization and calibration of $\mathbf{W}$ is modeled by adding a small update term, $\delta\mathbf{W}$, such that $\widehat{\mathbf{W}} = \mathbf{W} + \delta\mathbf{W}$. To accurately quantize the weights \emph{after training}, it is necessary to define a loss function to measure the quantization error. In the response-agnostic setting \citep{obc,optq,quip,dettmers2024spqr}, the quantization error, $\varepsilon_{\ell_2}$, is measured by the $\ell_2$ loss between the output of the quantized and original layer
\begin{equation}
        \varepsilon_{\ell_2}
        =
         \mathbb{E}_{\mathbf{x}^{(l)}} \Bigr[  \| \mathbf{W} \mathbf{x}^{(l)} - \widehat{\mathbf{W}} \mathbf{x}^{(l)} \|^2_2 \Bigr] \\
        = 
         \mathbb{E}_{\mathbf{x}^{(l)}} \Bigr[ \mathrm{tr} (\delta\mathbf{W} \mathbf{x}^{(l)} \mathbf{x}^{(l)^\top} \delta\mathbf{W}^\top) \Bigr] 
        = 
        \mathrm{tr} (\delta\mathbf{W} \bar{\mathbf{H}} {\delta\mathbf{W}}^\top),
\label{eq:rag-loss}\end{equation}
where $\bar{\mathbf{H}} = \mathbb{E}_{\mathbf{x}^{(l)}} [ \mathbf{x}^{(l)}\mathbf{x}^{(l)^\top} ]$ is the response-agnostic Hessian and it is assumed that the rows of the weight matrix independently contribute to $\varepsilon_{\ell_2}$ \citep{obc}.

Next, an iterative calibration procedure is performed on the columns of $\mathbf{W}$ to minimize the quantization error \citep{obc,optq,dettmers2024spqr}. At the $q^{\mathrm{th}}$ iteration, $\mathbf{W}_{:,q}$ is quantized to $\widehat{\mathbf{W}}_{:,q}$, and the remaining columns are updated to minimize $\varepsilon_{\ell_2}$. Thus, the problem is formulated as the following optimization 
\begin{argmini}
    {\delta\mathbf{W}}{ \mathrm{tr} (\delta\mathbf{W} \bar{\mathbf{H}} \delta\mathbf{W}^\top),}
    {}{}
    \addConstraint{\delta\mathbf{W}_{:,q}}{=\widehat{\mathbf{W}}_{:,q} - \mathbf{W}_{:,q}}.
    \label{eq:quadratic_setting}
\end{argmini}
Having solved the optimization problem \eqref{eq:quadratic_setting} according to \citep{obc,optq}, the optimal update at the $q^{\mathrm{th}}$ iteration is
\begin{equation}
\label{eq:update_w}
    \delta\mathbf{W}^*_q = - \frac{\mathbf{W}_{:,q} - \widehat{\mathbf{W}}_{:,q}}{[\bar{\mathbf{H}}^{-1}]_{q,q}} [\bar{\mathbf{H}}^{-1}]_{q,:}.
\end{equation}

The most salient weights are labeled as outliers. However, various outlier mitigation techniques have been developed \citep{dettmers2024spqr,squeezellm,billm,gholami2022survey}. Equation \eqref{eq:sensitivity} is used to measure the saliency of each weight
\begin{equation}
    \label{eq:sensitivity}
    s_{j,k} = \frac{(\mathbf{W}_{j,k} - \widehat{\mathbf{W}}_{j,k})^2}{[\bar{\mathbf{H}}^{-1}]_{k,k}} .
\end{equation}

\section{Proposed Output-adaptive Calibration}
% \subsection{Output-adaptive Calibration}
\label{sec:OAC}
In contrast to the setting presented in Section \ref{sec:agnostic}, we propose calibrating the weights toward minimizing the difference between the model output loss before and after quantization, hence the reason for the name \textbf{Output-adaptive Calibration (OAC)}. In other words, we measure the quantization error based on the distortion of the cross-entropy loss i.e. $\mathcal{L}_{\mathrm{CE}}(\mathbf{x}, \mathbf{y}, {\ttheta})$ after quantizing the weights, where $\ttheta=\mathrm{vec}[\mathbf W^{(l)}]_{l=1}^L$. The quantization is denoted by $\widehat{\ttheta} = \ttheta + \delta\ttheta $ in which $\delta\ttheta$ is the update term. Therefore, the output-adaptive error is
\begin{equation}
\label{eq:exact-error}
\begin{split}
    \varepsilon_{\mathrm{OAC}}
    &=
    \mathbb{E}_{(\mathbf{x},\mathbf{y})} [\mathcal{L}_{\mathrm{CE}}(\mathbf{x}, \mathbf{y}, \widehat{\ttheta}) - \mathcal{L}_{\mathrm{CE}}(\mathbf{x}, \mathbf{y}, \ttheta)] \\
    &= \delta\ttheta^\top \bar{\mathbf{g}} + \delta\ttheta^\top \bar{\mathbf{H}}^{(\ttheta)}_\mathrm{OAC} \delta\ttheta + \mathcal{O}(\|\delta\ttheta\|^3),
\end{split}
\end{equation}
where $\bar{\mathbf{H}}^{(\ttheta)}_\mathrm{OAC} = \mathbb{E}_{(\mathbf{x}, \mathbf{y})} \Bigr[ \frac{\partial^2 \mathcal{L}_{\mathrm{CE}}}{\partial \ttheta \partial \ttheta^\top}  \Bigr] \in \mathbb{R}^{D \times D}$ is the output-adaptive Hessian of the entire weights and $\bar{\mathbf{g}} = \mathbb{E}_{(\mathbf{x}, \mathbf{y})} \Bigr[ \frac{\partial \mathcal{L}_{\mathrm{CE}}}{\partial \ttheta} \Bigr] \simeq 0$  since the model is trained \citep{adaround}. For LLMs, the computation of $\bar{\mathbf{H}}^{(\ttheta)}_\mathrm{OAC}$ is infeasible due to its huge size $\mathcal{O}(D^2)$. In Sections \ref{sec:cross-layer-ind}, \ref{sec:cross-row-ind}, and \ref{sec:aggregate}, we illustrate how our proposed approach circumvents the computation of exact $\bar{\mathbf{H}}^{(\ttheta)}_\mathrm{OAC}$.

\subsection{Cross-layer Independence} \label{sec:cross-layer-ind}
Following the layer-wise quantization and calibration recipe of our proposed method, we assume that the linear layers are independent and their output-adaptive Hessians can be computed independently. Therefore, $\bar{\mathbf{H}}^{(\ttheta)}_\mathrm{OAC}$ is approximated by a block-diagonal matrix, where the $l^{\mathrm{th}}$ non-zero diagonal block, denoted by $\bar{\mathbf{H}}^{(l)}_\mathrm{OAC}$, corresponds to the output-adaptive Hessian of the $l^{\mathrm{th}}$ linear layer, as shown in Figure \ref{fig:Hessian-states}:1). To further simplify the notation, we drop $l$ from $\bar{\mathbf{H}}^{(l)}_\mathrm{OAC}$ when there is no risk of confusion. Having assumed the cross-layer independence, the quantization error during the calibration of each linear layer is computed as
\begin{equation}
\label{eq:cross-layer-error}
    \varepsilon_{\mathrm{OAC}}
    =
    \mathbb{E}_{(\mathbf{x},\mathbf{y})} [\mathcal{L}_{\mathrm{CE}}(\mathbf{x}, \mathbf{y}, \widehat{\ttheta}) - \mathcal{L}_{\mathrm{CE}}(\mathbf{x}, \mathbf{y}, \ttheta)] \\
    \simeq 
    \delta\mathbf{w}^\top \bar{\mathbf{H}}_\mathrm{OAC} \delta\mathbf{w},
\end{equation}
where $\mathbf{w} = \mathrm{vec}(\mathbf{W})$, $\bar{\mathbf{H}}_{\mathrm{OAC}} = \mathbb{E}_{(\mathbf{x},\mathbf{y})} \Bigr[ \frac{\partial^2 \mathcal{L}_{\mathrm{CE}}}{\partial \mathbf{w} \partial \mathbf{w}^\top}  \Bigr] \in \mathbb{R}^{d_\mathrm{row}d_\mathrm{col} \times d_\mathrm{row}d_\mathrm{col}}$ is the output-adaptive Hessian of the linear layer, and $\delta\mathbf{w} = \mathrm{vec}(\widehat{\mathbf{W}} - \mathbf{W})$ is the update term.

Considering the dimensions of linear layers in LLMs, $\bar{\mathbf{H}}_\mathrm{OAC}$ is a huge matrix $\mathcal{O}(d_\mathrm{row}^2 d_\mathrm{col}^2)$ with memory-intensive computation. Some existing methods, such as \citep{squeezellm,brecq}, ignore all of the inter-element correlations and assume that all of the weights are independent. Such methods approximate the Hessian of each layer by a diagonal matrix comprising the diagonal elements of the Fisher information matrix. However, this over-restrictive assumption leads to accuracy drop \citep{obs}. To achieve a more accurate Hessian approximation, we relax this assumption as described in Section \ref{sec:cross-row-ind}. 

\subsection{Cross-row Independence} \label{sec:cross-row-ind}
\begin{figure}[!t]
    \centering
    \includegraphics[width=\textwidth]{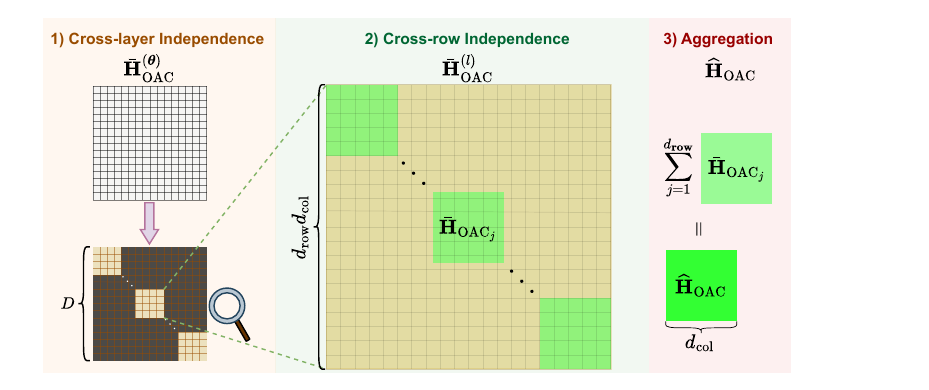}
    \caption{This figure shows our proposed steps to reduce the computational complexity of the output-adaptive Hessian. 1) The Hessian of each linear layer is independently computed. 2) The Hessian of each linear layer becomes block diagonal according to the rows independence assumption. 3) All of the row-wise Hessians are aggregated to reduce the memory footprint.}
    \label{fig:Hessian-states}
\end{figure}

In a linear layer, the output elements are independently generated by the inner product of the rows of the weight matrix and the input columns. In our proposed method, we assume that only the rows of $\mathbf{W}$ are independent. Consequently, each row of $\mathbf{W}$ has its own row-wise output-adaptive Hessian that is computed separately. Therefore, $\bar{\mathbf{H}}_\mathrm{OAC}$ is approximated by a block diagonal matrix, where the $j^{\mathrm{th}}$ block, $\bar{\mathbf{H}}_{\mathrm{OAC}_j}$, corresponds to the Hessian of $\mathbf{W}_{j,:}$ as shown in Figure \ref{fig:Hessian-states}:2). 

The rows of $\mathbf{W}$ are assumed to be independent and the columns of $\mathbf{W}$ are iteratively calibrated \citep{obs,obc,optq,dettmers2024spqr}, therefore, the optimal update is found by solving
\begin{argmini}
    {\delta\mathbf{W}}
    {\varepsilon_{\mathrm{OAC}} 
    \simeq 
    \delta\mathbf{W}^\top \bar{\mathbf{H}}_\mathrm{OAC} \delta\mathbf{W}
    \simeq
    \sum_{j=1}^{d_\mathrm{row}} {\delta\mathbf{W}_{j,:} \bar{\mathbf{H}}_{\mathrm{OAC}_j} \delta\mathbf{W}_{j,:}^\top}
    }
    {}{}
    \addConstraint{\delta\mathbf{W}_{:,q}}{=\widehat{\mathbf{W}}_{:,q} - \mathbf{W}_{:,q}}
    \label{appeq:loss3},
\end{argmini}
where only the row-wise Hessians, $\bar{\mathbf{H}}_{\mathrm{OAC}_j}$, are needed. Therefore, the memory footprint is reduced from $\mathcal{O}(d_\mathrm{row}^2 d_\mathrm{col}^2)$ to $\mathcal{O}(d_\mathrm{row} d_\mathrm{col}^2)$. However, $\mathcal{O}(d_\mathrm{row} d_\mathrm{col}^2)$ is still quite huge compared to the output-agnostic $\ell_2$ Hessian $\mathcal{O}(d_\mathrm{col}^2)$ that is used in the PTQ baselines \citep{optq}. We then aggregate the row-wise Hessians to further reduce the required memory as described in Section~\ref{sec:aggregate}. 

\subsection{Aggregation of Row-wise Hessians}
\label{sec:aggregate}
Computation of the row-wise Hessians $\bar{\mathbf{H}}_{\mathrm{OAC}_j}$ used in equation \eqref{appeq:loss3} is associated with a large memory footprint. We propose replacing $\bar{\mathbf{H}}_{\mathrm{OAC}_j}$ with $\widehat{\mathbf{H}}_\mathrm{OAC} = \sum_{j=1}^{d_\mathrm{row}} \bar{\mathbf{H}}_{\mathrm{OAC}_j}$ to mitigate the memory complexity. Thus, our simplified optimization problem transforms to

\begin{argmini}
    {\delta\mathbf{W}}{\varepsilon_{\mathrm{OAC}} 
    =   \mathbb{E}_{(\mathbf{x},\mathbf{y})} [\mathcal{L}_{\mathrm{CE}}(\mathbf{x}, \mathbf{y}, \widehat{\ttheta}) - \mathcal{L}_{\mathrm{CE}}(\mathbf{x}, \mathbf{y}, \ttheta)] 
    \simeq \mathrm{tr} (\delta\mathbf{W} \widehat{\mathbf{H}}_\mathrm{OAC} \delta\mathbf{W}^\top),}
    {}{}
    \addConstraint{\delta\mathbf{W}_{:,q}}{=\widehat{\mathbf{W}}_{:,q} - \mathbf{W}_{:,q}}.
    \label{eq:rad_quadratic_setting}
\end{argmini}

Note that $\mathrm{tr} (\delta\mathbf{W} \widehat{\mathbf{H}}_\mathrm{OAC} \delta\mathbf{W}^\top)$ is an upper bound for $\sum_{j=1}^{d_\mathrm{row}} {\delta\mathbf{W}_{j,:} \bar{\mathbf{H}}_{\mathrm{OAC}_j} \delta\mathbf{W}_{j,:}^\top}$ since all $\bar{\mathbf{H}}_{\mathrm{OAC}_j}$ matrices are positive definite. Hence, equation \eqref{eq:rad_quadratic_setting} approximates \eqref{appeq:loss3} as supported empirically in our experiments.

% As equation \eqref{eq:rad_quadratic_setting} shows, the output-adaptive optimization problem is similar to the output-agnostic case. Therefore, by replacing $\widehat{\mathbf{H}}_{\mathrm{OAC}}$ in equation \eqref{eq:update_w}, the update term is computed in the output-adaptive scenario. Also, $\widehat{\mathbf{H}}_{\mathrm{OAC}}$ can be used in equation \eqref{eq:sensitivity} to detect the outliers considering the cross-entropy loss instead of the layer-wise quantization error. The only remaining challenge is computing the output-adaptive Hessian, which will be addressed in the next section.

\subsection{Approximation of the Output-adaptive Hessian}
To further clarify our proposed output-adaptive approach, we start this section by computing the output-adaptive Hessian for the weights of a logistic regression classifier. We then generalize our approach to the linear layers of LLMs.

\subsubsection{$\bar{\mathbf{H}}_\mathrm{OAC}$ for a Binomial Logistic Regression Classifier}
\label{sec:logistic_regression}
Let $\mathbf{w} \in \mathbb{R}^d$ be the weights of a logistic regression model trained using cross-entropy. $\mathbf{x}_i \in \mathbb{R}^d$ is the input, and $y_i \in \{0,1\}$ is its label. Equation \eqref{eq:logistic_regression_p} shows the underlying mechanism of the model.
\begin{equation}
\label{eq:logistic_regression_p}
    P_\mathbf{w}(y_i=1 | \mathbf{x}_i) = \pi_\mathbf{w}(\mathbf{x}_i) = \frac{e^{\mathbf{w}^\top \mathbf{x}_i}}{1 + e^{\mathbf{w}^\top \mathbf{x}_i}}
\end{equation}

Equations \eqref{eq:logistic_regression_g} and \eqref{eq:logistic_regression_H} show the gradient and the Hessian obtained for the $i^{\mathrm{th}}$ sample, respectively,
\begin{equation}
    \label{eq:logistic_regression_g}
    \mathbf{g}[i]
    \coloneq
    \frac{\partial \mathcal{L}_{\mathrm{CE}}(\mathbf{w}; \mathbf{x}_i, y_i)}{\partial \mathbf{w}}
    =
    \mathbf{x}_i \Bigr[ \pi_\mathbf{w}(\mathbf{x}_i) - y_i \Bigr],
\end{equation}

\begin{equation}
    \label{eq:logistic_regression_H}
    \frac{\partial^2 \mathcal{L}_{\mathrm{CE}}(\mathbf{w}; \mathbf{x}_i) }{\partial \mathbf{w} \partial \mathbf{w}^\top}
    = \mathbf{x}_i \Bigr[ \pi_\mathbf{w}(\mathbf{x}_i) [1 - \pi_\mathbf{w}(\mathbf{x}_i)] \Bigr] \mathbf{x}_i^\top.
\end{equation}

Based on the Fisher information identity, we approximate \footnote{The proof is provided in Appendix \ref{app:logistic_regression}.} the expected Hessian over $N$ samples as
\begin{align}
    \label{eq:logistic_regression_H_g}
    \bar{\mathbf{H}}_{\mathrm{OAC}}
    &= \nonumber
    \mathbb{E}_{(\mathbf{x}_i, y_i)} \Biggr[ \frac{\partial^2 \mathcal{L}_{\mathrm{CE}}(\mathbf{w}; \mathbf{x}_i) } {\partial \mathbf{w} \partial \mathbf{w}^\top} \Biggr]
    =     
    \mathbb{E}_{(\mathbf{x}_i, y_i)} \Biggr[ \frac{\partial \mathcal{L}_{\mathrm{CE}}(\mathbf{w}; \mathbf{x}_i, y_i)}{\partial \mathbf{w}} \frac{\partial \mathcal{L}_{\mathrm{CE}}(\mathbf{w}; \mathbf{x}_i, y_i)^\top}{\partial \mathbf{w}} \Biggr]  \\\nonumber
    &= 
    \mathbb{E}_{\mathbf{x}_i} \Biggr[ \mathbf{x}_i \mathbb{E}_{y_i | \mathbf{x}_i} \Bigr[ [\pi_\mathbf{w}(\mathbf{x}_i) - y_i ]^2 \Bigr] \mathbf{x}_i^\top \Biggr]
    =
    \mathbb{E}_{\mathbf{x}_i} \Biggr[ \mathbf{x}_i \Bigr[ \pi_\mathbf{w}(\mathbf{x}_i) [1 - \pi_\mathbf{w}(\mathbf{x}_i)] \Bigr] \mathbf{x}_i^\top \Biggr]  \\ 
    &\simeq 
    \frac{1}{N} \sum_{i=1}^{N} {\mathbf{g}[i] \mathbf{g}[i]^\top}.
\end{align}

% \bar{\mathbf{H}}_{\mathrm{OAC}}
%     &= \nonumber
%     \mathbb{E}_{(\mathbf{x}_i, y_i)} \left[ \frac{\partial^2 \mathcal{L}_{\mathrm{CE}}(\mathbf{w}; \mathbf{x}_i) } {\partial \mathbf{w} \partial \mathbf{w}^\top} \right]
%     = 
%     \mathbb{E}_{(\mathbf{x}_i, y_i)} \left[ \frac{\partial \mathcal{L}_{\mathrm{CE}}(\mathbf{w}; \mathbf{x}_i, y_i)}{\partial \mathbf{w}} \frac{\partial \mathcal{L}_{\mathrm{CE}}(\mathbf{w}; \mathbf{x}_i, y_i)^\top}{\partial \mathbf{w}} \right]  \\ \nonumber
%      &=
%      \mathbb{E}_{\mathbf{x}_i} \left[ \mathbf{x}_i  \mathbb{E}_{y_i | \mathbf{x}_i} \left[ \left(\pi_\mathbf{w}(\mathbf{x}_i) - y_i \right)^2 \right] \mathbf{x}_i^\top \right] 
%      =
%      \mathbb{E}_{\mathbf{x}_i} \left[ \mathbf{x}_i \left[ \pi_\mathbf{w}(\mathbf{x}_i) \left[1 - \pi_\mathbf{w}(\mathbf{x}_i)\right] \right]  \mathbf{x}_i^\top \right]

As shown in equation \eqref{eq:logistic_regression_H_g}, the term $y_i$ appears when approximating the second derivative with $\mathbf{g}[i] \mathbf{g}[i]^\top$ based on the Fisher information identity. Hence, we call our method output-adaptive.

\subsubsection{Generalization of $\widehat{\mathbf{H}}_\mathrm{OAC}$ for Linear Layers of LLMs}
\label{sec:gencase}
% After approximating the output-adaptive Hessian for a vector of correlated weights, we generalize our approach to compute $\widehat{{\mathbf{H}}}_{\mathrm{OAC}}$ for linear layers of LLMs. 
Having established the computation of $\bar{\mathbf{H}}_\mathrm{OAC}$ for a binomial logistic regression classifier, we propose to generalize our approach to compute $\widehat{{\mathbf{H}}}_{\mathrm{OAC}}$ for the linear layers of LLMs as follows. Let $\mathbf{W} \in \mathbb{R}^{d_\mathrm{row} \times d_\mathrm{col}}$ be the weight matrix of the $l^{\mathrm{th}}$ linear layer in an LLM resulted by concatenating $d_\mathrm{row}$ rows. Following the cross-layer independence assumption, the Hessian of $\mathbf{W}$ is independent of other layers. Also, based on the cross-row independence assumption, the Hessian for each row of $\mathbf{W}$ is independent of other rows. Using equation \eqref{eq:logistic_regression_H_g}, the output-adaptive Hessian of the $j^\mathrm{th}$ row is approximated by equation \eqref{eq:row-wiseH}, where $\mathbf{G}_{j,:}[i]$ is the $j^\mathrm{th}$ row of the gradient matrix, $\mathbf{G}[i] \in \mathbb{R}^{d_\mathrm{row} \times d_\mathrm{col}}$, computed for the $i^\mathrm{th}$ calibration sample.
\begin{equation}
    \label{eq:row-wiseH}
\displaystyle
\bar{\mathbf{H}}_{\mathrm{OAC}_j} \simeq \frac{1}{N} \sum_{i=1}^{N} {\mathbf{G}_{j,:}^\top[i] \mathbf{G}_{j,:}[i]}.
\end{equation}

Then, all of the row-wise Hessians are aggregated to compute $\widehat{\mathbf{H}}_{\mathrm{OAC}}$. However, to optimize the computations, equation \eqref{eq:fullRAH} is used to directly compute the aggregation of all row-wise Hessians without individually computing them

\begin{equation}
    \label{eq:fullRAH}
    \begin{split}
    \displaystyle
    \widehat{\mathbf{H}}_{\mathrm{OAC}} \simeq \sum_{j=1}^{d_\mathrm{row}} \bar{\mathbf{H}}_{\mathrm{OAC}_j} 
    &\simeq
    \frac{1}{N} \sum_{i=1}^{N} {\sum_{j=1}^{d_\mathrm{row}} {\mathbf{G}_{j,:}[i]^\top \mathbf{G}_{j,:}[i]}} \\
    &= \frac{1}{N} \sum_{i=1}^{N} {\mathbf{G}[i]^\top \mathbf{G}[i]}.
    \end{split}
        % = \sum_{j=1}^{d_\mathrm{row}} { \frac{1}{N} \sum_{i=1}^{N} {\mathbf{g}[i,j] \mathbf{g}^\top[i,j]}} \\
\end{equation}

\section{OAC Pipeline}
\label{sec:pipeline}

\begin{algorithm}[!t]
\caption{OAC Pipeline}
\label{alg:oac}
\begin{algorithmic}[1]
\REQUIRE model, $N$ data samples\\
% \STATE $N \coloneq$ len(data)
% \STATE freeze(model) \\
\FOR{$j = 1$ \TO model.num\_layers}
    \STATE $\text{block} \coloneq \text{model.block}[j]$ \\
    \#\# Phase 1: Computation of  $\widehat{\mathbf{H}}_\mathrm{OAC}$ \#\# \\
    % \STATE unfreeze(block) \\
    \FOR{$i=1$ \TO $N$}
    \STATE loss = Cross-Entropy(model(data[$i$]), data[$i$]) \\
    \STATE loss.backward()
        \FOR{layer $\in$ block.linear\_layers}
            \STATE $\mathbf{W} \coloneq$ layer.weight
            \STATE $\mathbf{G}[i] = \sfrac{\partial \text{loss}}{\partial \mathbf{W}}$
            \STATE $\widehat{\mathbf{H}}_\text{OAC} += \frac{1}{N} \mathbf{G}^\top[i] \mathbf{G}[i]$
        \ENDFOR
    \ENDFOR \\
    \#\# Phase 2: Calibration by $\widehat{\mathbf{H}}_\mathrm{OAC}$ \#\# \\
    \FOR{layer $\in$ block.linear\_layers}
        \STATE $\widehat{\mathbf{W}} = \text{Hessian-based Calibration} (\mathbf{W}, \widehat{\mathbf{H}}_\mathrm{OAC}) $
    \ENDFOR
\ENDFOR
% \For{\textnormal{layer} $\in$ \textnormal{linear_layers(block)}}
% }
\end{algorithmic}

\end{algorithm}

Our proposed method, OAC, comprises two main phases for PTQ of LLMs weights. (i) Computing the output-adaptive Hessian of each weight matrix. (ii) Calibrating each weight matrix using its output-adaptive Hessian. To develop a complete PTQ pipeline, we integrate a Hessian-based calibration technique with our proposed method. The OAC pipeline is described in Algorithm \ref{alg:oac}.

\paragraph{Computation of $\widehat{\mathbf{H}}_\mathrm{OAC}$} 
OAC uses equation \eqref{eq:fullRAH} to approximate the Hessian of each weight matrix. Therefore, it requires computing the gradients of each weight matrix for all calibration samples. We propose computing the gradients of the linear layers inside each transformer block simultaneously, while other transformer blocks are frozen. This avoids repeating the costly output generation and backpropagation for all linear layers. As described in  \ref{alg:oac}, the transformer blocks are iteratively selected. For each block, the model outputs are generated using the calibration samples. 
%After measuring the cross-entropy loss, backpropagation is performed to compute the gradients of the linear layers inside the block.
Following the calculation of the cross-entropy loss, OAC employs backpropagation to compute the gradients of the weights associated with the linear layers within the block.
Then, the $\widehat{\mathbf{H}}_\mathrm{OAC}$ of the linear layers within the block are separately approximated based on equation \eqref{eq:fullRAH} and are used in the next phase of our proposed PTQ method.

\begin{figure}[!h]
    \centering
    \includegraphics[width=\textwidth]{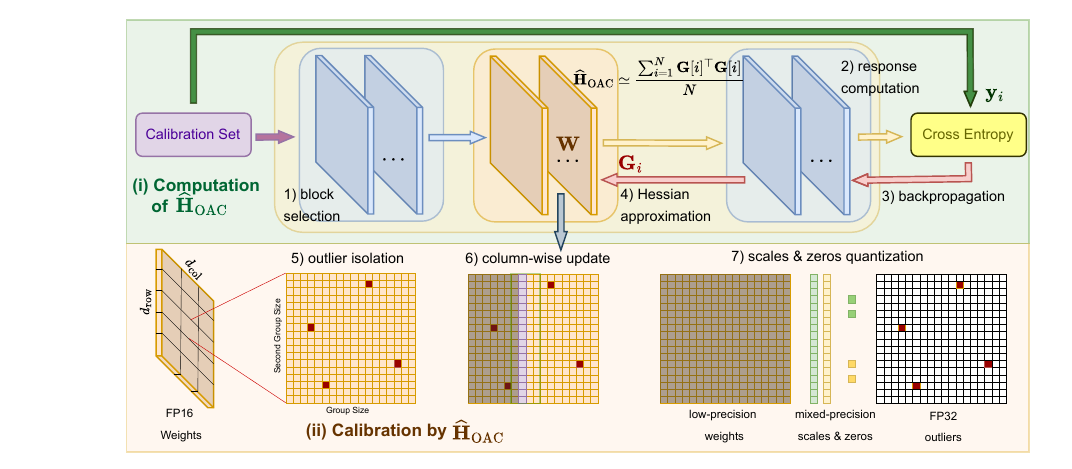}
    \caption{A demonstration of the OAC steps for 2-bit PTQ of LLMs. 1) The transformer blocks are iteratively selected for calibration. 2) The final outputs for the calibration samples are generated. 3) The generated outputs are compared with the ground truth outputs to compute the loss and gradients. 4) The output-adaptive Hessians of linear layers inside the block are approximated. 5) For each linear layer, the outliers are detected and isolated using $\widehat{\mathbf{H}}_\mathrm{OAC}$. 6) Column-wise calibration is performed to reduce the quantization error. 7) The quantization scales and zeros go through a second round of quantization to reduce the average bit width. Steps 5, 6, and 7 are integrated into our method from \citep{dettmers2024spqr}.}
    \label{fig:algorithm}
\end{figure}

\paragraph{Calibration by $\widehat{\mathbf{H}}_\mathrm{OAC}$}
To develop OAC for accurate 2-bit PTQ of LLMs, the following steps from SpQR \citep{dettmers2024spqr} are integrated into our method. The salient weights are detected and isolated using equation \eqref{eq:sensitivity} while $\bar{\mathbf{H}}$ is replaced by our proposed output-adaptive Hessian $\widehat{\mathbf{H}}_{\mathrm{OAC}}$.
Then, the column-wise updating process is performed to reduce the distortion of the cross-entropy loss. Likewise, the update term is computed by replacing $\widehat{\mathbf{H}}_{\mathrm{OAC}}$ in equation \eqref{eq:update_w}. Finally, the scales and zeros are quantized to reduce the average bit hence enabling smaller group quantization and keeping more outliers in the FP32 format. Figure \ref{fig:algorithm} illustrates the complete steps of OAC, implemented for 2-bit PTQ of LLMs. Furthermore, to show the effectiveness of our proposed output-adaptive Hessian computation in binary PTQ, we integrated the $\widehat{\mathbf{H}}_{\mathrm{OAC}}$ to the calibration procedure of BiLLM \citep[Algorithm 1]{billm}. Experimental results presented in Section \ref{sec:experimental} show that our proposed approach surpasses both BiLLM and SpQR in the PTQ of LLMs.

\section{Experiments and Results}\label{sec:experimental}
This section provides the key experimental results in addition to a summary of the settings used in our experiments. We refer to the appendix for further discussion of the findings, hyper-parameters, and settings. 

\subsection{Experimental Settings}
We investigate various language models with different sizes including OPT \citep{opt}, LLaMa \citep{llama}, and LLaMa 2 \citep{llama2} families. The calibration set comprises 128 sequences of 2048 tokens. To evaluate the performance of the quantized models on language modeling tasks, we report their \textit{perplexity} on C4 \citep{c4} and WikiText2 \citep{merity2017pointer}. Also, Language Model Evaluation Harness (LMEH) \citep{eval-harness} is utilized for evaluating the reasoning abilities of the quantized models. We report the zero-shot accuracy on WinoGrande \citep{winogrande}, PiQA \citep{piqa}, HellaSwag \citep{hellaswag}, ARC-easy, and ARC-challenge \citep{arc} in addition to the five-shot exact match on GSM8K \citep{gsm8k} datasets. See Appendix \ref{app:datasets} for more details on the datasets.

In our experimental results, we compare our method with the latest state-of-the-art PTQ methods that are developed for LLMs including Round to Nearest (RTN) \citep{llm-int8}, OPTQ \citep{optq}, QuIP \citep{quip}, and SpQR \citep{dettmers2024spqr} on 2- and 3- bit quantization in addition to OmniQuant \citep{shao2024omniquant} as an efficient QAT method. SqueezeLLM \citep{squeezellm} is also included in the 3-bit quantization experiments. Moreover, BiLLM \citep{billm} is selected as the most recent baseline for binary PTQ. Detailed information on the experimental settings such as the quantization configurations are provided in the Appendix \ref{app:config}. 

\subsection{Results}
Quantization of LLMs becomes more complicated and exhibits a larger accuracy degradation as (i) the model size decreases, and (ii) the overall average bit width to accommodate lower precision weights, fewer mitigated outliers, and larger group quantization decreases. As such, we investigate 2-bit PTQ\footnote{See Appendix \ref{app:detailed_results} for the details of binary, 2-bit, and 3-bit PTQ results.} of small to large LLMs with 1.3B to 30B parameters. Table \ref{tab:results} compares the performance of 2-bit quantized LLaMa and OPT models on the language modeling and reasoning tasks. Our experimental results show that OAC significantly outperforms the state-of-the-art baselines.

To further evaluate our proposed method, OAC, at extreme low-precision quantization, we apply OAC to the binarization of the LLaMa family. Table \ref{tab:binary} shows the performance of OAC compared to BiLLM. Our experimental results show that OAC significantly outperforms BiLLM.  

The experimental results reveal the importance of output-adaptive calibration in low-precision quantization scenarios. Our proposed approach, OAC, achieves a better accuracy compared to the other methods, especially, in more complicated scenarios where the average bit width or model size is relatively smaller.

  \begin{table}[!t]
\centering
\caption{Comparison of the perplexity and reasoning score for 2-bit quantized LLaMa and OPT models. The "LMEH" column shows the average score over the LMEH reasoning tasks.}
\label{tab:results}
\resizebox{\textwidth}{!}{%
  \begin{tabular}{ccccc|ccccc}
 \toprule
   Method & Avg Bits & C4  & WikiText2  & LMEH & Method & Avg Bits & C4 & WikiText2 & LMEH \\
   \midrule
   \multicolumn{5}{c|}{LLaMa2-7B} & \multicolumn{5}{c}{LLaMa2-13B}  \\
  \midrule
   Baseline & 16 & 7.03 & 5.47 & 56.19  & Baseline & 16 & 6.50 & 4.88 & 60.34 \\
   RTN & 2.25 & 4.6e3 & 4.3e3 & 29.37 & RTN  & 2.25 & 1.4e2 & 1.2e2 & 32.36 \\
   OPTQ & 2.25 & 1.5e2 & 2.0e2 & 30.46 & OPTQ & 2.25 & 52.34 & 47.09 & 31.81  \\
   OmniQuant & 2.25 & 15.74 & 11.16 & 39.34 & OmniQuant & 2.25 & 11.62 & 8.31 & 44.95 \\
   QuIP &  2   & 51.22 & 67.92 & 32.06 & QuIP &   2  & 11.63 & 9.81 & 44.78 \\
   SpQR & 2.09 & 13.22 & 11.09 & 42.90 & SpQR & 2.09 & 9.81 & 7.58 & 49.29 \\
   OAC (ours) & 2.09 & \textbf{11.9}0 & \textbf{9.48} & \textbf{45.56} & OAC (ours) & 2.09 & \textbf{9.49} & \textbf{7.39} & \textbf{49.96} \\
  \midrule
  \multicolumn{5}{c|}{LLaMa-13B} & \multicolumn{5}{c}{LLaMa-30B}  \\
  \midrule
  Baseline & 16 & 6.61 & 5.09 & 58.86 & Baseline & 16 & 5.95 & 4.10 & 64.39 \\
  RTN  & 2.25 & 4.5e2 & 7.8e2 & 31.40 & RTN  & 2.25 & 99.23 & 68.06 & 33.23 \\
  OPTQ & 2.25 & 24.56 & 19.16 & 30.88 & OPTQ & 2.25 & 11.07 & 8.63  & 46.29 \\
  OmniQuant & 2.25 & 10.70 & 7.78 & 47.03 & OmniQuant & 2.25 & 9.69 & 6.98 & 49.04 \\
  QuIP &  2   & 10.95 & 9.30 & 46.11 & QuIP &   2  & 9.28 & 7.48 & 49.51 \\
  SpQR & 2.09 & 9.61 & 7.63 & 48.35 & SpQR & 2.09 & 8.25 & \textbf{6.29} & 54.16 \\
  OAC (ours) & 2.09 & \textbf{9.45} & \textbf{7.45} & \textbf{50.13} & OAC (ours) & 2.09 & \textbf{8.22} & 6.31 & \textbf{54.97} \\
  \midrule
  \multicolumn{5}{c|}{OPT-13B} & \multicolumn{5}{c}{OPT-30B}  \\
  \midrule
  Baseline & 16 & 11.54 & 10.13 & 58.72 & Baseline & 16 & 10.91 & 9.56 & 60.97 \\
  RTN  & 2.25 & 2.7e4 & 7.6e4 & 34.85 & RTN  & 2.25 & 6.4e3 & 1.3e4 & 35.12 \\
  OPTQ & 2.25 & 15.67 & 15.62 & 50.14 & OPTQ & 2.25 & 13.43 & 12.85 & 54.05 \\
  OmniQuant & 2.25 & 20.47 & 15.70 & 50.71 & OmniQuant & 2.25 & 13.65 & 11.38 & 55.59 \\
  QuIP & 2  & 14.07 & 13.41 & 53.76 & QuIP & 2    & 12.41 & 11.27 & 57.20 \\
  SpQR & 2.09 & 13.28 & 12.16 & 55.07 & SpQR & 2.09 & 12.00 & 10.72 & 57.63 \\
  OAC (ours) & 2.10 & \textbf{13.25} & \textbf{11.75} & \textbf{55.86} & OAC (ours) & 2.09 & \textbf{11.99} & \textbf{10.48} & \textbf{58.20} \\
  \bottomrule
  \end{tabular}%
  }
\end{table}

\begin{table}[!t]
\centering
  \caption{Comparison of the perplexity and reasoning score for binarized LLaMa models. The "LMEH" column shows the average score over the LMEH reasoning tasks.}
  \label{tab:binary}
\resizebox{\textwidth}{!}{%
  \begin{tabular}{ccccc|ccccc}
 \toprule
   Method & Avg Bits & C4  & WikiText2  & LMEH & Method & Avg Bits & C4 & WikiText2 & LMEH \\
   \midrule
   \multicolumn{5}{c|}{LLaMa2-7B} & \multicolumn{5}{c}{LLaMa2-13B}  \\
  \midrule
   Baseline & 16 & 7.03 & 5.47 & 56.19  & Baseline & 16 & 6.50 & 4.88 & 60.34 \\
   % SpQR & 1.13 & NaN & NaN & 28.65 & SpQR & 1.13 & 7.0e4 & 7.1e4 & 29.35 \\
   BiLLM & 1.08 & 28.00 & 25.59 & 35.61 & BiLLM & 1.08 & 23.06 & 18.54 & 37.01 \\
   OAC (ours) & 1.09 & \textbf{21.64} & \textbf{19.50} & \textbf{38.74} & OAC (ours) & 1.08 & \textbf{15.25} & \textbf{13.15} & \textbf{42.42} \\
  \midrule
  \multicolumn{5}{c|}{LLaMa-7B} & \multicolumn{5}{c}{LLaMa-13B}  \\
  \midrule
  Baseline & 16 & 7.11 & 5.68 & 55.15 & Baseline & 16 & 6.61 & 5.09 & 58.86 \\
  % SpQR & 1.13 & 1.9e5 & 1.7e5 & 29.28 & SpQR & 1.13 & 8.3e4 & 8.5e4 & 29.04 \\
  BiLLM & 1.09 & 27.82 & 30.73 & 35.88 & BiLLM & 1.09 & \textbf{14.93} & 14.13 & 42.65 \\
  OAC (ours) & 1.09 & \textbf{19.82} & \textbf{17.79} & \textbf{38.84} &  OAC (ours) & 1.09 & 15.17 & \textbf{14.05} & \textbf{44.31} \\
  \bottomrule
  \end{tabular}%
  }
\end{table}

\section{Conclusion}

We proposed a novel Output-adaptive Calibration (OAC) approach to incorporate the model output in the PTQ of LLMs. Our proposed approach minimizes the quantization error based on the distortion of the output cross-entropy loss, leading to a more accurate quantized model. To reduce the computational complexity of OAC, we introduced an approximation to the output-adaptive Hessian based on the Fisher information identity. We also delved into the details of deploying our approximated Hessian to update the weight matrices and identify the salient weights within the PTQ pipeline. We compared OAC with several PTQ methods on various LLMs, and found OAC to be more accurate on various language modeling and reasoning tasks. Our experimental results demonstrate that our proposed PTQ method outperforms the state-of-the-art by a significant margin in extreme low-precision (e.g. 2-bit and binary).

%%%%%%%%%%%%%%%%%%%%%%%%%%%%%%%%%%%%%%%%%%%%%%%%%%%%%%%%%%%%
\bibliographystyle{plainnat}
\bibliography{main}

\appendix

% \hl{Hyper-parameters, hardware setting, random seed,}

\section{Response-adaptive Hessian for Logistic Regression}
\label{app:logistic_regression}
Following Section \ref{sec:logistic_regression}, let $\mathbf{w} \in \mathbb{R}^d$ represent the weights of a binomial logistic regression classifier. Given the input $\mathbf{x}_i \in \mathbb{R}^d$, the model predicts the label $y_i \in \{0,1\}$. Note that equation \eqref{appeq:logistic_regression_p} shows the model formulation. Also, equations \eqref{appeq:logistic_regression_loss}, \eqref{appeq:logistic_regression_g}, and \eqref{appeq:logistic_regression_H} show the cross-entropy loss, gradient, and Hessian computed for each sample, respectively.
\begin{equation}
\label{appeq:logistic_regression_p}
    P_\mathbf{w} (y_i=1 | \mathbf{x}_i) = \pi_\mathbf{w}(\mathbf{x}_i) = \frac{e^{\mathbf{w}^\top \mathbf{x}_i}}{1 + e^{\mathbf{w}^\top \mathbf{x}_i}}
\end{equation}
\begin{equation}
\label{appeq:logistic_regression_loss}
    \mathcal{L}_{CE}(\mathbf{w}; \mathbf{x}_i, y_i) = - \Bigr[ y_i \log \pi_\mathbf{w}(\mathbf{x}_i) + [1-y_i] \log [1 - \pi_\mathbf{w}(\mathbf{x}_i)] \Bigr]
\end{equation}
\begin{equation}
    \label{appeq:logistic_regression_g}
    \mathbf{g}[i]
    \coloneq
    \frac{\partial \mathcal{L}_{\mathrm{CE}}(\mathbf{w}; \mathbf{x}_i, y_i)}{\partial \mathbf{w}}
    =
    \mathbf{x}_i \Bigr[ \pi_\mathbf{w}(\mathbf{x}_i) - y_i \Bigr]
\end{equation}

\begin{equation}
    \label{appeq:logistic_regression_H}
    \mathbf{H}[i]
    \coloneq
    \frac{\partial^2 \mathcal{L}_{\mathrm{CE}}(\mathbf{w}; \mathbf{x}_i) }{\partial \mathbf{w} \partial \mathbf{w}^\top}
    = \mathbf{x}_i \Bigr[ \pi_\mathbf{w}(\mathbf{x}_i) [1 - \pi_\mathbf{w}(\mathbf{x}_i)] \Bigr] \mathbf{x}_i^\top
\end{equation}

The expected output-adaptive Hessian can be approximated using the gradients as follows,
\begin{equation}
\label{appeq:logistic_regression_H_g}
\begin{split}
    \mathbb{E}_{(\mathbf{x}_i, y_i)} \Bigr[ \mathbf{g}[i] \mathbf{g}[i]^\top \Bigr] 
    &=
    \mathbb{E}_{(\mathbf{x}_i, y_i)} \Biggr[ \frac{\partial \mathcal{L}_{\mathrm{CE}}(\mathbf{w}; \mathbf{x}_i, y_i)}{\partial \mathbf{w}} \frac{\partial \mathcal{L}_{\mathrm{CE}}(\mathbf{w}; \mathbf{x}_i, y_i)^\top}{\partial \mathbf{w}} \Biggr] \\
    &=
    \mathbb{E}_{\mathbf{x}_i} \Biggr[ \mathbb{E}_{y_i|\mathbf{x}_i} \Biggr[ \mathbf{x}_i \Bigr[ \pi_\mathbf{w}(\mathbf{x}_i) - y_i \Bigr]^2  \mathbf{x}_i^\top \Biggr] \Biggr] \\
    &=
    \mathbb{E}_{\mathbf{x}_i} \Biggr[ \mathbf{x}_i  \Biggr[ \mathbb{E}_{y_i|\mathbf{x}_i}  \Bigr[ \pi_\mathbf{w}(\mathbf{x}_i) - y_i \Bigr]^2  \Biggr]  \mathbf{x}_i^\top \Biggr] \\
    &=
    \mathbb{E}_{\mathbf{x}_i} \Biggr[ \mathbf{x}_i \Biggr[ \mathbb{E}_{y_i|\mathbf{x}_i}  \Bigr[ \pi^2_\mathbf{w}(\mathbf{x}_i) + y^2_i - 2y_i\pi_\mathbf{w}(\mathbf{x}_i) \Bigr] \Biggr] \mathbf{x}_i^\top \Biggr] \\
    &=
    \mathbb{E}_{\mathbf{x}_i} \Biggr[ \mathbf{x}_i \Biggr[ \mathbb{E}_{y_i|\mathbf{x}_i}  \Bigr[ \pi^2_\mathbf{w}(\mathbf{x}_i) \Bigr] + \mathbb{E}_{y_i|\mathbf{x}_i}  \Bigr[ y^2_i \Bigr] - \mathbb{E}_{y_i|\mathbf{x}_i}  \Bigr[ 2y_i\pi_\mathbf{w}(\mathbf{x}_i) \Bigr] \Biggr] \mathbf{x}_i^\top \Biggr] \\
    &=
    \mathbb{E}_{\mathbf{x}_i} \Biggr[ \mathbf{x}_i \Bigr[ \pi^2_\mathbf{w}(\mathbf{x}_i)  + \pi_\mathbf{w}(\mathbf{x}_i) - 2\pi^2_\mathbf{w}(\mathbf{x}_i) \Bigr] \mathbf{x}_i^\top \Biggr] \\
    &=
    \mathbb{E}_{\mathbf{x}_i} \Biggr[ \mathbf{x}_i \Bigr[ \pi_\mathbf{w}(\mathbf{x}_i) [1 - \pi_\mathbf{w}(\mathbf{x}_i)] \Bigr] \mathbf{x}_i^\top \Biggr] \\
    &=
    \mathbb{E}_{(\mathbf{x}_i, y_i)} \Biggr[ \mathbf{x}_i \Bigr[ \pi_\mathbf{w}(\mathbf{x}_i) [1 - \pi_\mathbf{w}(\mathbf{x}_i)] \Bigr] \mathbf{x}_i^\top \Biggr] \\
    &=
    \mathbb{E}_{(\mathbf{x}_i, y_i)} \Biggr[ \frac{\partial^2 \mathcal{L}_{\mathrm{CE}}(\mathbf{w}; \mathbf{x}_i) }{\partial \mathbf{w} \partial \mathbf{w}^\top} \Biggr] \\
    &=
    \mathbb{E}_{(\mathbf{x}_i, y_i)} \Bigr[ \mathbf{H}[i] \Bigr].
\end{split}
\end{equation}

Since the calibration set contains $N$ samples, the output-adaptive Hessian is empirically estimated by 
\begin{equation}
\label{appeq:empirical_estimation}
    \bar{\mathbf{H}}_\mathrm{OAC} 
    =
    \mathbb{E}_{(\mathbf{x}_i, y_i)} \Bigr[ \mathbf{H}[i] \Bigr]
    =
    \mathbb{E}_{(\mathbf{x}_i, y_i)} \Bigr[ \mathbf{g}[i] \mathbf{g}[i]^\top \Bigr]
    \simeq
    \frac{1}{N} \sum_{i=1}^{N} {\mathbf{g}[i] \mathbf{g}[i] ^\top}.
\end{equation}

\section{Aggregation of Row-wise Hessians}
% Equation \eqref{eq:fullRAH} shows how $\widehat{\mathbf{H}}_\mathrm{OAC}$ is approximated by directly computing the aggregation of the row-wise Hessians, without individually computing the row-wise Hessians. Figure \ref{fig:multirow} demonstrates how Equation \eqref{eq:fullRAH} works for further clarification.

\begin{figure}[!h]
    \centering
    \includegraphics{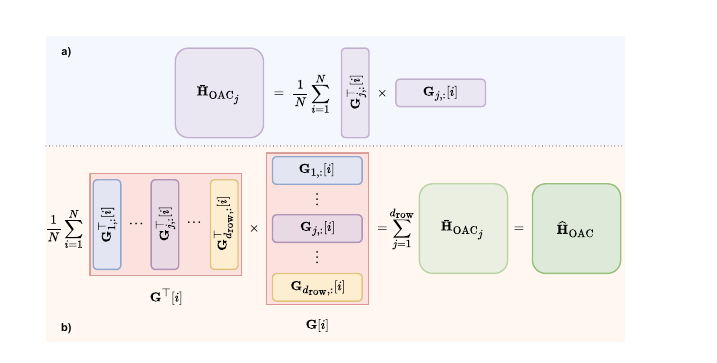}
    \caption{A demonstration of equation \eqref{eq:fullRAH} in  Section \ref{sec:aggregate}, where  a) shows the Hessian of each row and, b) shows the aggregation of row-wise Hessians.}
    \label{fig:multirow}
\end{figure}

Figure \ref{fig:multirow} further clarifies the equation \eqref{eq:fullRAH} in Section \ref{sec:aggregate}.  We inserted this figure here due to limited space in the main article.

\section{Numerical Stability of OAC}
\label{app:stability}
% We investigate three aspects of the numerical stability of our method as follows.
To ensure the robustness of our proposed output-adaptive Hessian, we analyze its numerical stability from three perspectives as follows.

\subsection{Half-precision Gradient  Computation} 
\label{app:precision}
In our experiments, both forward and backward passes, including the gradient computations are performed in the single-precision floating-point (FP32). However, to significantly reduce the time and memory footprint, the gradient computations can be performed in the half-precision floating-point (FP16). Table \ref{tab:precision} compares using FP16 versus FP32 data types for the gradient computations of OAC to quantize LLaMa-7B and LLaMa-13B to 2 bits. In the FP16 mode, we scale the cross-entropy loss by $\{16, 32, 128, 256, 512, 1024\}$ and report the average and standard deviation. 

Our experimental results show that FP16 gradient computations reduce the time and memory footprint by about $64\%$ and $30\%$ respectively while maintaining the perplexity. Moreover, the perplexity results for the FP16 mode are robust to the loss scaling since they have a low standard deviation.

\begin{table}[!ht]
\centering
\caption{Comparison of the WikiText2 perplexity, time, and GPU memory footprint of OAC when the gradient computations are performed in FP32 and FP16.}
\label{tab:precision}
\begin{tabular}{ccccccc}
\toprule
       Model & \begin{tabular}[c]{@{}c@{}} Average \\ Bits \end{tabular} & \begin{tabular}[c]{@{}c@{}} Gradient \\ Computations Type \end{tabular} & \begin{tabular}[c]{@{}c@{}} Time \\ (hours:minutes) \end{tabular} & \begin{tabular}[c]{@{}c@{}} GPU Memory \\ (GB) \end{tabular} & \begin{tabular}[c]{@{}c@{}} WikiText2 \\ Perplexity \end{tabular} \\
       \midrule
        \multirow{2}{*}{LLaMa-13B} & \multirow{2}{*} {2.09} & FP32 & 8:50 & 146.12 & \textbf{7.45} \\
        % \cmidrule(r){3-6}
        & & FP16 & \textbf{3:03} & \textbf{100.28} & 7.50$_{\pm 0.04}$  \\
        \midrule
        \multirow{2}{*}{LLaMa-7B} & \multirow{2}{*} {2.09} & FP32 & 4:13 & 80.99 & 9.57 \\
        % \cmidrule(r){3-6}
        & & FP16 & \textbf{1:29} & \textbf{58.46} & \textbf{9.40}$_{\pm 0.07}$\\
        \bottomrule
\end{tabular}
\end{table}

\subsection{Hessian Regularization}
\label{app:H_regularization}
When employing the Hessian inverse for the calibration in PTQ, a regularization term is added to the diagonal elements of the Hessian matrix. This additional term serves to mitigate numerical instability, which is common in computing the inverse of matrices \citep{optq,quip,dettmers2024spqr,billm}.
\begin{equation}
\label{appeq:alpha}
\bar{\mathbf{H}} = \bar{\mathbf{H}} + \mathrm{diag}(\alpha *  \mathrm{mean}(\mathrm{diag}(\bar{\mathbf{H}})))     
\end{equation}
In our experiments, we tune $\alpha$ for all applicable methods as a hyperparameter by performing a greed search over $\{0.001, 0.01, 0.1, 1\}$ and selecting the one that achieves the best validation perplexity. Table \ref{tab:alpha} shows how tuning $\alpha$ affects the perplexity of the quantized LLaMa-7B.

\begin{table}[ht!]
    \centering
     \caption{Comparison of WikiText2 perplexity for the quantized LLaMa-7B when $\alpha \in \{0.001, 0.01, 0.1, 1 \}$. }
    \label{tab:alpha}
    \resizebox{\textwidth}{!}{%
    \begin{tabular}{ccccc|cccccc}
    \toprule
       $\alpha$ &  0.001 & 0.01 & 0.1 & 1 & $\alpha$ &  0.001 & 0.01 & 0.1 & 1 \\
        \midrule
        SpQR (2.09-bit) & 11.73 & 12.08 & 11.20 & \textbf{10.59}  & BiLLM (1.09-bit) & 293.29 & 44.11 & \textbf{30.73} & 43.73 \\
        \midrule
        OAC (2.09-bit)  & 11.67 & 11.28 & 10.17 & \textbf{9.57} & OAC (1.09-bit) & 28.79 & 19.81 & \textbf{17.79} & 19.19 \\
        \bottomrule
    \end{tabular}%
    }
\end{table}

\subsection{Hessian Reduction Over Calibration Samples}
\label{app:H_reduction}
As discussed previously, the approximation of $\widehat{\mathbf{H}}_\mathrm{OAC}$ for each linear layer is shown in equation \eqref{eq:fullRAH} where the output-adaptive Hessian is averaged over the calibration samples.
% Equation \eqref{eq:fullRAH} shows the $\widehat{\mathbf{H}}_\mathrm{OAC}$ approximation formula for each linear layer where the output-adaptive Hessian is averaged over the calibration samples. 
However, due to the small magnitude of the gradients computed over each calibration sample, we skip the division by $N$ to achieve better numerical stability in our experiments. Therefore, we use equation \eqref{appeq:reduction} to approximate the output-adaptive Hessian of each layer in our experiments.
\begin{equation}
    \label{appeq:reduction}
    \widehat{\mathbf{H}}_\mathrm{OAC} = \sum_{i=1}^{N} {\mathbf{G}^\top[i] \mathbf{G}[i]}
\end{equation}

Note that theoretically, scaling the Hessian matrix does not affect the calibration process. However, empirically, it may lead to slightly different results due to floating-point error induced by division. Table \ref{tab:reduction} shows the effect of the Hessian reduction technique on the WikiText2 perplexity of the 2-bit quantized LLaMa-13B. 

\begin{table}[!ht]
\centering
\caption{The effect of the Hessian reduction method on the perplexity of LLaMa-13B, quantized with OAC. "Mean" indicates using equation \eqref{eq:fullRAH} for the computation of the output-adaptive Hessian. "Sum" indicates skipping the division by $N$, as shown in equation \eqref{appeq:reduction}.}
\label{tab:reduction}
\begin{tabular}{cccccc}
\toprule
       Model & Avg Bits & Hessian Reduction  & C4 & WikiText2 \\
       \midrule
        \multirow{2}{*}{LLaMa-13B} & \multirow{2}{*} {2.09} & Mean & 9.49 & 7.58 \\
        % \cmidrule(r){3-5}
        & & Sum & \textbf{9.45} & \textbf{7.45} \\
        \bottomrule
\end{tabular}
\end{table}

\section{Seed Sensitivity}
\label{app:seed}
We compare the performance of our proposed method, OAC, to the most competitive baseline, SpQR, on 2-bit PTQ across four seeds, $\{0, 1376, 1997, 4695\}$, to verify the robustness of our results to the randomness. Table \ref{tab:seed} shows the mean and standard deviation of the perplexity and LMEH score of the 2-bit quantized OPT-13B. Our results show that OAC outperforms SpQR across the selected seeds. Note that due to the huge computational cost of performing all of the experiments with multiple seeds, we used one seed (seed=0) to produce other experimental results in this paper.

\begin{table}[!ht]
\centering
\caption{Comparison of the performance of OAC and SpQR on 2-bit PTQ for OPT-13B. The mean and standard deviation of the perplexity on C4, WikiText2, and PTB as well as the averaged reasoning score on LMEH tasks are reported across 4 seeds.}
\label{tab:seed}
\begin{tabular}{ccccccc}
\toprule
       Model & Avg Bits & Method & C4 & WikiText2 & PTB & LMEH \\
       \midrule
        \multirow{2}{*}{OPT-13B} & \multirow{2}{*} {2.09} & SpQR & 13.32$_{\pm 0.06}$ & 13.47$_{\pm 0.07}$ & 12.22$_{\pm 0.14}$ & 54.96$_{\pm 0.24}$ \\
        & & OAC & \textbf{13.25}$_{\pm 0.01}$ & \textbf{13.39}$_{\pm 0.09}$ & \textbf{11.81}$_{\pm 0.08}$ & \textbf{55.76}$_{\pm 0.20}$ \\
        \bottomrule
\end{tabular}
\end{table}

\section{Computational Cost of OAC}
\label{app:cost}
Our proposed method, OAC, approximates the output-adaptive Hessian of each weight matrix using the gradients of the layer, computed over the calibration samples. Therefore, OAC requires extra time, memory, and computational resources to compute the gradient matrices which is the limitation of OAC compared to the existing baselines for PTQ of LLMs. Table \ref{tab:cost} shows the computational costs of OAC compared to SpQR at 2-bit PTQ of the LLaMa models. Here, we mention a few points to further clarify the computational costs of our proposed method.

First, OAC computes the gradients over a limited calibration dataset to perform a Hessian-based calibration. In contrast, QAT methods often require larger datasets to learn parameters, which makes them more costly than our proposed method, OAC. 

Second, although OAC requires more time and memory compared to the existing response-agnostic methods such as SpQR, it can quantize LLMs within a reasonable time and memory budget with better accuracy. For example, it takes several hours to quantize mid-size LLMs using OAC. Also, the required memory to apply OAC on most LLMs is within the capacity of commonly used machines.

Third, as discussed in Section \ref{app:precision}, the half-precision (FP16) gradient computations significantly reduce the computational cost of OAC while maintaining accuracy.

\begin{table}[!ht]
\centering
\caption{Comparison of the required time and memory for applying OAC and SpQR at 2-bit PTQ of LLaMa-7B and LLaMa-13B. Also, the WikiText2 perplexity is provided in the last column.}
\label{tab:cost}
\begin{tabular}{ccccccc}
\toprule
        Model & \begin{tabular}[c]{@{}c@{}} Average \\ Bits \end{tabular} & Method & \begin{tabular}[c]{@{}c@{}} Time \\ (hours:minutes) \end{tabular} & \begin{tabular}[c]{@{}c@{}} GPU Memory \\ (GB) \end{tabular} & \begin{tabular}[c]{@{}c@{}} WikiText2 \\ Perplexity \end{tabular} \\
       \midrule
        \multirow{3}{*}{LLaMa-7B} & \multirow{3}{*} {2.09} & SpQR & 1:06 & 3.57 & 10.59 \\
        & & OAC$_{\mathrm{FP32}}$ & 4:13 & 80.99 & 9.57 \\
        & & OAC$_{\mathrm{FP16}}$ & 1:29 & 58.46 & 9.32 \\
        \midrule
        \multirow{3}{*}{LLaMa-13B} & \multirow{3}{*} {2.09} & SpQR & 1:37 & 5.62 & 7.63 \\
        & & OAC$_{\mathrm{FP32}}$ & 8:50 & 146.12 & 7.45  \\
        & & OAC$_{\mathrm{FP16}}$ & 3:03 & 100.28 & 7.50  \\
        \bottomrule
\end{tabular}
\end{table}

\section{Datasets}
\label{app:datasets}
\paragraph{Calibration Set} We create the calibration sets used by the studied PTQ methods by randomly selecting 128 samples of length 2048 from C4 \citep{c4}, RedPajama \citep{together2023redpajama}, or WikiText2 \citep{merity2017pointer} datasets. The C4 samples are mostly used for the calibration of OPT models. The LLaMa 1 and LLaMa 2 models are calibrated mostly based on the RedPajama samples. However, some PTQ baselines do not support using RadPajama in their official implementation for quantizing the LLaMa models. In those cases, C4 samples are used instead of the RadPajama samples for the calibration of such methods. Also, the WikiText2 samples are used in the OmniQuant experiments for calibrating the OPT and, LLaMa 1, and LLaMa 2 families. Refer to Section \ref{app:config} for the detailed implementation configurations of each studied method.   

\paragraph{Validation set} We randomly select 256 data samples from the C4 validation set to create the validation dataset that is used for tuning $\alpha$.

\paragraph{Test Datasets} The WikiText2 \citep{merity2017pointer} and PTB \citep{ptb} perplexities are reported on their standard test sets. The test dataset for reporting the C4 perplexity is created by randomly selecting 1024 data samples from the C4 validation set. We selected WinoGrande \citep{winogrande}, PiQA \citep{piqa}, HellaSwag \citep{hellaswag}, GSM8K \citep{gsm8k}, ARC-easy, and ARC-challenge \citep{arc} datasets from Language Model Evaluation Harness (LMEH) \citep{eval-harness} to evaluate the reasoning abilities of the quantized models. We emphasize that GSM8K is excluded from the OPT evaluations due to the poor performance of uncompressed OPT models on this challenging task while we keep GSM8K in the evaluation of the quantized LLaMa models.

\section{Quantization Configurations}
\label{app:config}
We use up to 8$\times$NVIDIA V100-32G GPUs for our experiments. Also, the Stochastic Gradient Descent (SGD) optimizer implemented by PyTorch \citep{torch} is used in our experiments to compute the gradients of the linear layers. We implemented our experiments using the Hugging Face transformers \citep{transformers} framework. Here, the exact experimental configurations are reported for the baseline PTQ methods that are studied in our experiments.

\paragraph{RTN} We implemented this method based on the SpQR official implementation\footnote{See \href{https://github.com/Vahe1994/SpQR}{https://github.com/Vahe1994/SpQR}.} for RTN \citep{llm-int8}. However, we integrated group quantization in our version of RTN to significantly improve the accuracy. RTN is applied in the 2- and 3-bit PTQ experiments and the quantization group size is set to 128.

\paragraph{OPTQ} This method is implemented using the SpQR official implementation. We set the quantization group size to 128 to apply OPTQ \citep{optq} in the 2- and 3-bit PTQ experiments. The calibration samples are drawn from C4 and RedPajama to quantize the OPT and LLaMa models, respectively. We also tune the Hessian regularization factor, $\alpha$ for this method.

\paragraph{OmniQuant} This method is implemented using its Offical implementation\footnote{See \href{https://github.com/OpenGVLab/OmniQuant}{https://github.com/OpenGVLab/OmniQuant}.} in 2- and 3-bit weight-only quantization experiments. As proposed in the paper \citep{shao2024omniquant}, 128 samples from WikiText2 are randomly selected to compose the calibration set and the quantization group size is set to 128. We reproduce the quantized OPT models using this method following the hyperparameters recommended in their paper. To employ OmniQuant for the quantization of the LLaMa 1 and LLaMa 2 models, we download and evaluate the quantized checkpoints provided in the official public repository.

\paragraph{SpQR} We use the SpQR official implementation to apply SpQR in the binary, 2-bit, and 3-bit PTQ experiments. Table \ref{tab:spqr_config} shows the hyperparameters that are used for running the SpQR \citep{dettmers2024spqr} experiments. We also tune the Hessian regularization factor, $\alpha$ for this method.

\begin{table}[!ht]
    \centering
    \caption{Configurations of the SpQR experiments.}
    \label{tab:spqr_config}
    \begin{tabular}{ccccc}
    \toprule
         \begin{tabular}[c]{@{}c@{}} Model \\ Family \end{tabular}  & \begin{tabular}[c]{@{}c@{}} Calibration \\ Set Source \end{tabular} & \begin{tabular}[c]{@{}c@{}} Group \\ Size \end{tabular} & \begin{tabular}[c]{@{}c@{}} Weights, Scales, \\  $\&$ Zeros Bits \end{tabular} & \begin{tabular}[c]{@{}c@{}} Outlier \\ Threshold \end{tabular}  \\
         \midrule
         \multirow{2}{*}{OPT} & \multirow{2}{*}{C4} & \multirow{2}{*}{64} & 2 & 3.5 \\
          & & & 3 & 0.75 \\         
         \midrule
         \multirow{3}{*}{LLaMa1$\&$2} & C4 & 32 & 1 & 20 \\
         & RedPajama & 64 & 2 & 3.5 \\
         & RedPajama & 64 & 3 & 0.65 \\
         \bottomrule
    \end{tabular}
\end{table}

\paragraph{QuIP} This method is implemented using its official implementation\footnote{See \href{https://github.com/Cornell-RelaxML/QuIP}{https://github.com/Cornell-RelaxML/QuIP}.}. We tune $\alpha$ and draw the calibration samples from C4 \citep{quip} in the 2- and 3-bit PTQ experiments.

\paragraph{SqueezeLLM} We use the quantized checkpoints provided by the SqueezeLLM \citep{squeezellm} official implementation\footnote{See \href{https://github.com/SqueezeAILab/SqueezeLLM}{https://github.com/SqueezeAILab/SqueezeLLM}.} to reproduce this method in the 3-bit PTQ experiments. 

\paragraph{BiLLM} This method \citep{billm} is applied for binary PTQ of the LLaMa 1 and LLaMA 2 models using its official implementation\footnote{See \href{https://github.com/Aaronhuang-778/BiLLM}{https://github.com/Aaronhuang-778/BiLLM}.}. We tune the Hessian regularization factor, $\alpha$ for this method. Also, the calibration set is selected from C4.

\paragraph{OAC} We set the gradient computation data type to FP32 To quantize the models with less than 13B parameters. However, for OPT-30B and LLaMa-30B, we set the gradient computation percision to FP16 due to memory limitation. As discussed in Section \ref{app:precision}, we tune the loss scale for FP16 mode. The optimal scales are 512 and 32 for OPT-30B and LLaMa-30B respectively. We also tune the Hessian regularization factor, $\alpha$ for this method. Table \ref{tab:oac_config} shows the hyperparameters that are used for running the OAC experiments.
\begin{table}[!ht]
    \centering
    \caption{Configurations of the OAC experiments.}
    \label{tab:oac_config}
    \begin{tabular}{cccccc}
    \toprule
         \begin{tabular}[c]{@{}c@{}} Model \\ Family \end{tabular} & \begin{tabular}[c]{@{}c@{}} Calibration \\ Set Source \end{tabular} & \begin{tabular}[c]{@{}c@{}} Group \\ Size \end{tabular} & \begin{tabular}[c]{@{}c@{}} Weights, Scales, \\  $\&$ Zeros Bits \end{tabular} & \begin{tabular}[c]{@{}c@{}} Calibration \\ Phase Source \end{tabular} & \begin{tabular}[c]{@{}c@{}} Outlier \\ Threshold \end{tabular}  \\
         \midrule
         \multirow{2}{*}{OPT} & \multirow{2}{*}{C4} & \multirow{2}{*}{64} & 2 & SpQR & 3.5 \\
          & & & 3 & SpQR & 0.75 \\         
         \midrule
         \multirow{3}{*}{LLaMa1$\&$2} & C4 & N/A & 1 (Weights only) & BiLLM & N/A \\
         & RedPajama & 64 & 2 & SpQR & 3.5 \\
         & RedPajama & 64 & 3 & SpQR & 0.65 \\
         \bottomrule
    \end{tabular}
\end{table}

\section{Detailed Experimental Results}
\label{app:detailed_results}
The detailed experimental results of applying our method as well as the studied baselines for binary, 2-bit, and 3-bit quantization of the OPT, LLaMa 1, and LLaMa 2 models are reported in Tables \ref{tab:fbinary}, \ref{tab:opt-2}, \ref{tab:llama-2}, and \ref{tab:3-bit}. The results show that our proposed method, OAC, significantly outperforms all other state-of-the-art baselines, especially at low precision PTQ such as binary and 2-bit quantizations. Although we have reported the performance of SpQR for the sake of completeness in Table \ref{tab:fbinary}, we note that SpQR is not designed for binary PTQ, hence the reason for the poor performance of SpQR in binary PTQ.

\begin{table}[!h]
\centering
\caption{Comparison of the perplexity and reasoning score for the binary LLaMa models.}
\label{tab:fbinary}
\resizebox{\textwidth}{!}{%
  \begin{tabular}{cccccccccccc}
  \toprule
  Method & Avg Bits & C4 $\downarrow$ & WikiText2 $\downarrow$ & WinoGrande $\uparrow$ & PiQA $\uparrow$ & HellaSwag $\uparrow$ & ARC-challenge $\uparrow$ & ARC-easy $\uparrow$ & GSM8K $\uparrow$ & Average $\uparrow$ \\
  \midrule
  \multicolumn{11}{c}{LLaMA-7B} \\
  \midrule
  Baseline & 16 & 7.11 & 5.68 & 69.77 & 78.67 & 56.97 & 41.89 & 75.25 & 8.34 & 55.15 \\
  SpQR & 1.13 & 1.9e5 & 1.7e5 & 50.75 & 51.69 & 25.5 & 22.35 & 25.38 & 0.0 & 29.28\\
  BiLLM & 1.09 & 27.82 & 30.73  & 56.59 & 61.64 & 32.87 & 20.31 & 43.9 & 0.0 & 35.88 \\
  OAC (ours) & 1.09 & \textbf{19.82} & \textbf{17.79} & \textbf{57.46} & \textbf{66.21} & \textbf{35.58} & \textbf{22.70} & \textbf{50.46} &\textbf{ 0.61} & \textbf{38.84} \\
  \midrule
  \multicolumn{11}{c}{LLaMA2-7B} \\
  \midrule
  Baseline & 16 & 7.03 & 5.47 & 68.98 & 78.07 & 57.14 & 43.43 & 76.26 & 13.27 & 56.19 \\
  SpQR & 1.13 & NaN & NaN & 49.57 & 49.51 & 25.04 & 22.7 & 25.08 & 0.0 & 28.65 \\
  BiLLM & 1.08 & 28.0 & 25.59 & 54.7 & 61.97 & 32.82 & 21.76 & 42.42 & 0.0 & 35.61 \\
  OAC (ours) & 1.08 & \textbf{21.64} & \textbf{19.50} & \textbf{57.30} & \textbf{65.83} & \textbf{34.55} & \textbf{23.12} & \textbf{50.80} & \textbf{0.83} & \textbf{38.74} \\
  \midrule
  \multicolumn{11}{c}{LLaMA-13B} \\
  \midrule
  Baseline & 16 & 6.61 & 5.09 & 72.69 & 79.16 & 59.93 & 46.42 & 77.36 & 17.59 & 58.86 \\
  SpQR & 1.13 & 8.3e4 & 8.5e4 & 49.09 & 51.52 & 25.49 & 23.46 & 24.66 & 0.0 & 29.04 \\
  BiLLM & 1.09 & \textbf{14.93} & 14.13 & \textbf{62.75} & 68.72 & \textbf{40.5}5 & 25.6 & 56.52 & \textbf{1.74} & 42.65 \\
  OAC (ours) & 1.09 & 15.17 & \textbf{14.05} & 62.35 & \textbf{71.06} & 40.32 & \textbf{28.67} & \textbf{61.78} & 1.67 & \textbf{44.31} \\
  \midrule
  \multicolumn{11}{c}{LLaMA2-13B} \\
  \midrule
  Baseline & 16 & 6.50 & 4.88 & 72.22 & 79.16 & 60.04 & 48.46 & 79.34 & 22.82 & 60.34 \\
  SpQR & 1.13 & 7.0e4 & 7.1e4 & 49.17 & 52.61 & 25.68 & 22.95 & 25.67 & 0.0 & 29.35 \\
  BiLLM & 1.08 & 23.06 & 18.54 & 56.59 & 62.57 & 32.19 & 21.42 & 48.44 & 0.83 & 37.01 \\
  OAC (ours) & 1.08 & \textbf{15.25} & \textbf{13.15} & \textbf{59.43} & \textbf{68.72} & \textbf{39.39} & \textbf{26.96} & \textbf{59.05} & \textbf{0.99} & \textbf{42.42} \\
  \bottomrule
  \end{tabular}%
  }
\end{table}

\begin{table}[!h]
\centering
\caption{Comparison of the perplexity and reasoning score for the 2-bit OPT models.}
\label{tab:opt-2}
\resizebox{\textwidth}{!}{%
  \begin{tabular}{ccccccccccc}
  \toprule
  Method & Avg Bits & C4 $\downarrow$ & WikiText2 $\downarrow$ & PTB $\downarrow$ & WinoGrande $\uparrow$ & PiQA $\uparrow$ & HellaSwag $\uparrow$ & ARC-challenge $\uparrow$ & ARC-easy $\uparrow$ & Average $\uparrow$ \\
  \midrule
  \multicolumn{11}{c}{OPT-1.3B} \\
  \midrule
  Baseline & 16 & 15.46 & 14.63 & 15.63 & 59.83 & 71.76 & 41.49 & 23.21 & 57.07 & 50.67 \\
  RTN  & 2.25 & 7814.09 & 13005.91 & 7123.62 & 50.51 & 53.16 & 25.71 & 20.65 & 25.34 & 35.07 \\
  OPTQ & 2.25 & 79.07 & 239.34 & 162.05 & 49.64 & 53.05 & 26.09 & 20.05 & 27.15 & 35.20 \\ 
  OmniQuant & 2.25 & 34.52 & 28.04 & 38.88 & 55.88 & 63.22 & 32.53 & 19.45 & 44.11 & 43.04 \\
  QuIP & 2    & 26.77 & 32.61 & 34.29 & 53.75 & 65.94 & 34.93 & 20.56 & 45.37 & 44.11 \\
  SpQR & 2.11 & 24.14 & 27.44 & 30.76 & 54.30 & 64.42 & 35.02 & \textbf{21.08} & 45.75 & 44.11 \\
  OAC (ours) & 2.13 & \textbf{21.15} & \textbf{22.76} & \textbf{24.40} & \textbf{57.14} & \textbf{68.99} & \textbf{36.95} & 20.99 & \textbf{51.35} & \textbf{47.08} \\
  \midrule
  \multicolumn{11}{c}{OPT-6.7B} \\
  \midrule
  Baseline & 16 & 12.19 & 10.86 & 12.17 & 65.35 & 76.28 & 50.51 & 30.46 & 65.57 & 57.63 \\
  RTN  & 2.25 & 5245.55 & 7824.03 & 5656.0 & 47.43 & 53.05 & 25.67 & 20.90 & 25.76 & 34.56 \\
  OPTQ & 2.25 & 24.72 & 35.73 & 27.35 & 50.20 & 55.01 & 26.67 & 18.09 & 30.81 & 36.16 \\
  OmniQuant & 2.25 & 20.67 & 16.43 & 21.22 & 58.25 & 69.04 & 40.35 & 22.78 & 56.36 & 49.36 \\
  QuIP & 2    & 16.08 & 16.05 & 16.50 & 58.56 & 72.14 & 44.51 & 26.11 & 58.96 & 52.06 \\
  SpQR & 2.09 & 14.90 & 14.18 & 15.07 & \textbf{62.83} & 72.96 & 45.40 & 26.88 & \textbf{59.97} & 53.61 \\
  OAC (ours) & 2.10 & \textbf{14.42} & \textbf{13.61} & \textbf{14.77} & 62.12 & \textbf{74.43} & \textbf{45.99} & \textbf{27.65} & \textbf{59.97} & \textbf{54.03} \\
  \midrule
  \multicolumn{11}{c}{OPT-13B} \\
  \midrule  
  Baseline & 16 & 11.54 & 10.13 & 11.47 & 65.19 & 75.84 & 52.46 & 33.02 & 67.09 & 58.72 \\
  RTN  & 2.25 & 27564.01 & 76324.21 & 29895.23 & 49.41 & 52.29 & 26.21 & 20.22 & 26.14 & 34.85 \\
  OPTQ & 2.25 & 15.67 & 15.62 & 16.47 & 59.12 & 71.33 & 42.91 & 24.57 & 52.78 & 50.14 \\
  OmniQuant & 2.25 & 20.47 & 15.70 & 25.09 & 58.09 & 70.02 & 42.04 & 25.6 & 57.79 & 50.71 \\
  QuIP & 2    & 14.07 & 13.41 & 14.39 & 62.27 & 72.03 & 46.23 & 28.67 & 59.60 & 53.76 \\
  SpQR & 2.09 & 13.28 & 12.16 & 13.57 & 62.59 & 73.23 & 47.53 & 29.44 & 62.54 & 55.07 \\
  OAC (ours) & 2.10 & \textbf{13.25} & \textbf{11.75} & \textbf{13.50} & \textbf{63.30} & \textbf{74.70} & \textbf{47.64} & \textbf{30.63} & \textbf{63.01} & \textbf{55.86} \\
  \midrule
  \multicolumn{11}{c}{OPT-30B} \\
  \midrule
  Baseline & 16 & 10.91 & 9.56 & 11.09 & 68.19 & 77.64 & 54.3 & 34.73 & 69.99 & 60.97 \\
  RTN  & 2.25 & 6429.17 & 13064.10 & 6147.53 & 51.62 & 51.58 & 26.30 & 20.56 & 25.55 & 35.12 \\
  OPTQ & 2.25 & 13.43 & 12.85 & 13.65 & 62.51 & 72.85 & 47.02 & 27.82 & 60.06 & 54.05 \\
  OmniQuant & 2.25 & 13.65 & 11.38 & 13.42 & 63.14 & 74.97 & 47.79 & 28.5 & 63.55 & 55.59 \\
  QuIP & 2    & 12.41 & 11.27 & 12.62 & 65.51 & 75.41 & 49.75 & 30.38 & 64.94 & 57.20 \\ 
  SpQR & 2.09 & 12.00 & 10.72 & 12.32 & 65.19 & \textbf{75.52} & 50.44 & 31.31 & 65.70 & 57.63 \\
  OAC (ours) & 2.09 & \textbf{11.99} & \textbf{10.48} & \textbf{12.22} & \textbf{65.51} & 75.46 & \textbf{50.72} & \textbf{32.34} & \textbf{66.96} & \textbf{58.20} \\
 \bottomrule
  \end{tabular}%
  }
\end{table}

\begin{table}[!h]
\centering
\caption{Comparison of the perplexity and reasoning score for the 2-bit LLaMa models.}
\label{tab:llama-2}
\resizebox{\textwidth}{!}{%
  \begin{tabular}{ccccccccccc}
 \toprule
  Method & Avg Bits & C4 $\downarrow$ & WikiText2 $\downarrow$ & WinoGrande $\uparrow$ & PiQA $\uparrow$ & HellaSwag $\uparrow$ & ARC-challenge $\uparrow$ & ARC-easy $\uparrow$ & GSM8K $\uparrow$ & Average $\uparrow$ \\
  \midrule
  \multicolumn{11}{c}{LLaMa-7B} \\
  \midrule
  Baseline & 16 & 7.11 & 5.68 & 69.77 & 78.67 & 56.97 & 41.89 & 75.25 & 8.34 & 55.15 \\
  RTN  & 2.25 & 1041.25 & 1875.96 & 48.54 & 54.52 & 27.41 & 21.08 & 28.24 & 0.0 & 29.97 \\
  OPTQ & 2.25 & 76.3 & 103.8 & 52.96 & 58.76 & 31.36 & 20.22 & 35.31 & 0.0 & 33.1  \\
  OmniQuant & 2.25 & 13.32 & 9.66 & 60.3 & 67.41 & 41.5 & 29.44 & \textbf{59.97} & \textbf{0.61} & 43.21 \\
  QuIP &  2   & 19.2 & 17.69 & 56.67 & 65.29 & 38.15 & 25.0 & 51.14 & 0.0 & 39.38 \\
  SpQR & 2.09 & 12.56 & 10.59 & 59.75 & 69.7 & 44.63 & 28.07 & 59.81 & 0.45 & 43.73 \\
  OAC (ours) & 2.09 & \textbf{11.57} & \textbf{9.57}  & \textbf{62.51} & \textbf{71.44} & \textbf{46.61} & \textbf{29.61} & 59.34 & 0.53 & \textbf{45.01} \\
  \midrule
  \multicolumn{10}{c}{LLaMa2-7B} \\
  \midrule
  Baseline & 16 & 7.03 & 5.47 & 68.98 & 78.07 & 57.14 & 43.43 & 76.26 & 13.27 & 56.19 \\
  RTN & 2.25 & 4624.37 & 4273.34 & 50.04 & 52.94 & 26.01 & 19.80 & 27.44 & 0.00 & 29.37 \\
  OPTQ & 2.25 & 147.46 & 201.80 & 50.91 & 54.73 & 27.32 & 18.69 & 31.10 & 0.00 & 30.46 \\
  OmniQuant & 2.25 & 15.74 & 11.16 & 56.12 & 65.29 & 39.81 & 23.89 & 50.72 & 0.23 & 39.34 \\
  QuIP &  2   & 51.22 & 67.92 & 52.17 & 57.56 & 28.60 & 20.14 & 33.88 & 0.00 & 32.06 \\
  SpQR & 2.09 & 13.22 & 11.09 & 60.46 & 69.31 & 42.48 & 26.62 & 57.74 & 0.76 & 42.90 \\
  OAC (ours) & 2.09 & \textbf{11.90} & \textbf{9.48} & \textbf{61.72} & \textbf{71.49} & \textbf{46.07} & \textbf{30.89} & \textbf{62.29} & \textbf{0.91} & \textbf{45.56} \\
  \midrule
  \multicolumn{10}{c}{LLaMa-13B} \\
  \midrule  
  Baseline & 16 & 6.61 & 5.09 & 72.69 & 79.16 & 59.93 & 46.42 & 77.36 & 17.59 & 58.86 \\
  RTN  & 2.25 & 453.5 & 781.32 & 48.62 & 56.64 & 29.48 & 20.73 & 32.95 & 0.0 & 31.4 \\
  OPTQ & 2.25 & 24.56 & 19.16  & 49.96 & 55.93 & 28.29 & 17.49 & 33.0 & 0.61 & 30.88 \\
  QuIP &  2   & 10.95 & 9.3 & 64.25 & 72.36 & 47.64 & 30.97 & 60.61 & 0.83 & 46.11 \\
  OmniQuant & 2.25 & 10.7 & 7.78 & 63.93 & 72.47 & 46.76 & 31.4 & 66.46 & 1.14 & 47.03 \\
  SpQR & 2.09 & 9.61 & 7.63 & 64.96 & 73.78 & 49.66 & 34.39 & 65.66 & 1.67 & 48.35  \\
  OAC (ours) & 2.09 & \textbf{9.45} & \textbf{7.45} & \textbf{66.85} & \textbf{75.08} & \textbf{51.34} & \textbf{36.35} & \textbf{69.02} & \textbf{2.12} & \textbf{50.13} \\
  \midrule
  \multicolumn{10}{c}{LLaMa2-13B} \\
  \midrule
  Baseline & 16 & 6.50 & 4.88 & 72.22 & 79.16 & 60.04 & 48.46 & 79.34 & 22.82 & 60.34 \\
  RTN  & 2.25 & 139.01 & 122.07 & 48.93 & 58.54 & 33.07 & 21.16 & 32.45 & 0.00 & 32.36 \\
  OPTQ & 2.25 & 52.34 & 47.09 & 52.17 & 55.33 & 29.68 & 19.37 & 34.18 & 0.15 & 31.81 \\
  OmniQuant & 2.25 & 11.62 & 8.31 & 56.75 & 70.29 & 45.51 & 31.83 & 63.59 & 1.74 & 44.95 \\
  QuIP &   2  & 11.63 & 9.81 & 62.12 & 69.04 & 44.55 & 31.06 & 61.07 & 0.83 & 44.78 \\
  SpQR & 2.09 & 9.81 & 7.58 & 64.96 & \textbf{73.29} & 50.31 & 35.24 & 67.26 & \textbf{4.70} & 49.29 \\
  OAC (ours) & 2.09 & \textbf{9.49} & \textbf{7.39} & \textbf{66.38} & \textbf{73.29} & \textbf{50.93} & \textbf{36.09} & \textbf{69.53} & 3.56 & \textbf{49.96} \\
  \midrule
  \multicolumn{10}{c}{LLaMa-30B} \\
  \midrule
  Baseline & 16 & 5.95 & 4.1 & 75.77 & 81.07 & 63.33 & 52.82 & 80.43 & 32.9 & 64.39 \\
  RTN  & 2.25 & 99.23 & 68.06 & 53.75 & 57.94 & 27.54 & 20.9 & 39.23 & 0.0 & 33.23 \\
  OPTQ & 2.25 & 11.07 & 8.63  & 64.17 & 71.22 & 46.94 & 32.68 & 62.75 & 0.0 & 46.29 \\
  OmniQuant & 2.25 & 9.69 & 6.98 & 65.11 & 72.09 & 50.51 & 34.64 & 69.02 & 2.88 & 49.04 \\
  QuIP &   2  & 9.28 & 7.48 & 66.3 & 74.76 & 52.28 & 34.56 & 67.72 & 1.44 & 49.51 \\
  SpQR & 2.09 & 8.25 & \textbf{6.29} & 70.09 & 77.26 & 54.88 & 40.78 & 73.74 & 8.19 & 54.16 \\
  OAC (ours) & 2.09 & \textbf{8.22} & 6.31 & \textbf{71.11} & \textbf{77.31} & \textbf{55.43} & \textbf{40.87} & \textbf{75.08} & \textbf{10.01} & \textbf{54.97} \\
 \bottomrule
  \end{tabular}%
  }
\end{table}

\begin{table}[!h]
\centering
\caption{Comparison of the perplexity and reasoning score for the 3-bit LLaMa and OPT models.}
\label{tab:3-bit}
\resizebox{\textwidth}{!}{%
  \begin{tabular}{cccc|cccc|cccc}
  \toprule
  Method & Avg Bits & C4 $\downarrow$ & PTB $\downarrow$ & Method & Avg Bits & C4 $\downarrow$ & WikiText2 $\downarrow$ & Method & Avg Bits & C4 $\downarrow$ & WikiText2 $\downarrow$ \\
  \midrule
  \multicolumn{4}{c|}{OPT-6.7B} & \multicolumn{4}{c|}{LLaMa-7B} & \multicolumn{4}{c}{LLaMa2-7B} \\
  \midrule
  Baseline & 16 & 12.19 & 12.17 & Baseline & 16 & 7.11 & 5.68 & Baseline & 16 & 7.03 & 5.47 \\
  RTN  & 3.25 & 34.19 & 35.42 & RTN  & 3.25 & 8.85 & 7.01 & RTN & 3.25 & 8.64 & 6.66 \\
  OPTQ & 3.25 & 12.93 & 13.26 & OPTQ & 3.25 & 8.07 & 6.35 & OPTQ & 3.25 & 8.07 & 6.22 \\
  OmniQuant & 3.25 & 13.03 & 13.03 & OmniQuant & 3.25 & 7.85 & 6.14 & OmniQuant & 3.25 & 7.92 & 6.05 \\
  QuIP & 3    & 12.62 & 12.58 & QuIP &   3  & 8.73 & 6.96 & QuIP &   3  & 15.97 & 13.90 \\
  SqueezeLLM & 3.26 & 12.67 & 12.64 & SqueezeLLM & 3.24 & 7.64 & 6.13 & SqueezeLLM & 3 & 7.88 & 6.18  \\
  SpQR & 3.21 & 12.49 & 12.51 & SpQR & 3.21 & 7.62 & \textbf{6.05} & SpQR & 3.21 & 7.54 & 5.86 \\
  OAC (ours) & 3.22 & \textbf{12.44} & \textbf{12.43} & OAC (ours) & 3.21 & \textbf{7.51} & \textbf{6.05} & OAC (ours) & 3.21 & \textbf{7.48} & \textbf{5.83} \\
  \midrule
  \multicolumn{4}{c|}{OPT-13B} & \multicolumn{4}{c|}{LLaMa-13B} & \multicolumn{4}{c}{LLaMa2-13B} \\
  \midrule 
  Baseline & 16 & 11.54 & 11.47 & Baseline & 16 & 6.61 & 5.09 & Baseline & 16 & 6.50 & 4.88 \\
  RTN  & 3.25 & 46.22 & 57.38 & RTN & 3.25 & 7.58 & 5.88 & RTN & 3.25 & 7.30 & 5.52 \\
  OPTQ & 3.25 & 11.88 & 11.83 & OPTQ & 3.25 & 7.12 & 5.58 & OPTQ & 3.25 & 7.08 & 5.33 \\
  OmniQuant & 3.25 & 12.27 & 12.23 & OmniQuant & 3.25 & 7.09 & 5.44 & OmniQuant & 3.25 & 7.07 & 5.29 \\
  QuIP & 3    & 11.88 & 11.82 & QuIP &   3  & 7.15 & 5.62 & QuIP &   3  & 7.17 & 5.51 \\
  SqueezeLLM  & 3.26  & 11.96 & 11.93 & SqueezeLLM & 3.24 & 6.95 & 5.45 & SqueezeLLM & 3 & 7.04 & 5.36 \\
  SpQR & 3.2  & 11.75 & 11.73 & SpQR & 3.21 & 6.90 & 5.36 & SpQR & 3.21 & 6.81 & \textbf{5.13} \\
  OAC (ours) & 3.21 & \textbf{11.74} & \textbf{11.72} & OAC (ours) & 3.21 & \textbf{6.88} & \textbf{5.34} & OAC (ours) & 3.21 & \textbf{6.79} & 5.15 \\
  \bottomrule
  \end{tabular}%
  }
\end{table}

\section{Choice of the Hessian-based Calibration Method}
\label{app:calibration_method}
As mentioned in Section \ref{sec:pipeline}, different Hessian-based calibration methods including but not limited to OPTQ \citep{optq}, QuIP (and QuIP\#) \citep{quip,quip2}, SpQR \citep{dettmers2024spqr}, and BiLLM \citep{billm}, can be integrated into our proposed method to improve the accuracy of the quantized model. In the paper, we deployed SpQR and BiLLM as the Hessian-based calibration method to achieve state-of-the-art results for 2-bit and binary PTQ of LLMs (Refer to Tables \ref{tab:results} and \ref{tab:binary}). For example, Figure \ref{fig:algorithm} shows the steps of integrating SpQR into our method,  OAC$_{\text{SpQR}}$. 

In this section, we perform an ablation study to investigate if OAC consistently improves the accuracy if integrated with all of the studied Hessian-based calibration methods.

Table \ref{tab:calib_methods} shows the perplexity and LMEH average score of LLaMa-7B and LLaMa2-7B when quantized using the vanilla Hessian-based Calibration methods compared to integrating these methods to OAC. 

Based on our results, presented in Table \ref{tab:calib_methods}, integrating each Hessian-based calibration method into our proposed method, OAC, significantly improves the performance. These results also show the strength of our proposed Hessian computation approach. 

\begin{table*}[!t]
    \centering
      \begin{tabular}{cccccc}
     \toprule
      Model & Method & Avg Bits & C4 $\downarrow$ & WikiText2 $\downarrow$ & LMEH $\uparrow$ \\
      \midrule
      \multirow{8}{*}{LLaMa-7B} & OPTQ & 2.25 & 76.30 & 103.80 & 33.10 \\
      & OAC$_{\text{OPTQ}}$ & 2.25 & \textbf{49.21} & \textbf{50.95} & \textbf{34.62} \\
      \cmidrule(lr){2-6}
      & QuIP & 2 & 19.2 & 17.69 & 39.38 \\
      & OAC$_{\text{QuIP}}$ & 2 & \textbf{17.87} & \textbf{16.38} & \textbf{39.48} \\
      \cmidrule(lr){2-6}
      & SpQR & 2.09 & 12.56 & 10.59 & 43.73 \\
      & OAC$_{\text{SpQR}}$ & 2.09 & \textbf{11.57} & \textbf{9.57} & \textbf{45.01} \\
      \cmidrule(lr){2-6}
      & BiLLM & 1.09 & 27.82 & 30.73 & 35.88 \\
      & OAC$_{\text{BiLLM}}$ & 1.09 & \textbf{19.82} & \textbf{17.79} & \textbf{38.84} \\
      \midrule
      \multirow{8}{*}{LLaMa2-7B} & OPTQ & 2.25 & 147.46 & 201.8 & 30.46 \\
      & OAC$_{\text{OPTQ}}$ & 2.25 & \textbf{37.64} & \textbf{40.43} & \textbf{33.54} \\
      \cmidrule(lr){2-6}
      & QuIP & 2 & 51.22 & 67.92 & 32.06 \\
      & OAC$_{\text{QuIP}}$ & 2 & \textbf{28.41} & \textbf{28.35} & \textbf{35.35} \\
      \cmidrule(lr){2-6}
      & SpQR & 2.09 & 13.22 & 11.09 & 42.90 \\
      & OAC$_{\text{SpQR}}$ & 2.09 & \textbf{11.90} & \textbf{9.48} & \textbf{45.56} \\
      \cmidrule(lr){2-6}
      & BiLLM & 1.08 & 28.00 & 25.59 & 35.61 \\
      & OAC$_{\text{BiLLM}}$ & 1.09 & \textbf{21.64} & \textbf{19.50} & \textbf{38.74} \\
      \bottomrule
    \end{tabular}
    \caption{Comparison of the perplexity and reasoning score for the quantized LLaMa-7B and LLaMa2-7B models when the different Hessian-based calibration methods are integrated into OAC.}
  \label{tab:calib_methods}
  \end{table*}

\end{document}